\documentclass[10pt, two column, two side]{IEEEtran}

\usepackage{amssymb}
\usepackage{amsmath}
\usepackage{mathrsfs}
\usepackage{cite}
\usepackage{float}
\usepackage{bm}
\usepackage[linesnumbered, lined,boxed,commentsnumbered, ruled]{algorithm2e}
\usepackage{subfig}
\usepackage{graphicx,booktabs,multirow}
\usepackage{color}
\usepackage[colorlinks, linkcolor=color1, anchorcolor=blue, citecolor=color1]{hyperref}
\definecolor{color1}{RGB}{128,0,0}

\usepackage{amsthm}
\theoremstyle{definition}
\newtheorem{lemma}{Lemma}
\newtheorem{theorem}{Theorem}

\newtheorem{definition}{Definition}

\newcommand{\transpose}{\mathsf{T}}

\newcommand{\card}[1]{\left|{#1}\right|}
\newcommand{\norm}[1]{\left\|{#1}\right\|_{2}}

\newcommand{\Fnorm}[1]{\left\|{#1}\right\|_{F}}
\newcommand{\bracket}[1]{\left({#1}\right)}
\newcommand{\mbracket}[1]{\left[{#1}\right]}

\SetKwProg{Foreach}{for each }{ do }{end}
\SetKwProg{Foreachpara}{for each }{ do in parallel}{end}
\SetKwProg{Forallpara}{for all }{ do in parallel}{end}

\allowdisplaybreaks[4]

\title{Brain-Inspired Decentralized Satellite Learning in Space Computing Power Networks}
\author{Peng Yang, \textit{Graduate Student Member, IEEE}, Ting Wang, \textit{Senior Member, IEEE}, Haibin Cai,\\
Yuanming Shi, \textit{Senior Member, IEEE}, Chunxiao Jiang, \textit{Fellow, IEEE}, and Linling Kuang, \textit{Member, IEEE}
\thanks{P. Yang, T. Wang and H. Cai are with the MoE Engineering Research Center of Software/Hardware Co-design Technology and Application, the Shanghai Key Lab. of Trustworthy Computing, East China Normal University, Shanghai 200062, China (e-mail: 51205902030@stu.ecnu.edu.cn, twang@sei.ecnu.edu.cn, hbcai@sei.ecnu.edu.cn). Y. Shi is with the School of Information Science and Technology, ShanghaiTech University, Shanghai 201210, China (e-mail: shiym@shanghaitech.edu.cn). Chunxiao Jiang and Linling Kuang are with the Beijing National Research Center for Information Science and Technology, Tsinghua University, Beijing, 100084, China (e-mail: {jchx, kll}@tsinghua.edu.cn).}}

\begin{document}

\maketitle

\begin{abstract}
Satellite networks are able to collect massive space information with advanced remote sensing technologies, which is essential for real-time applications such as natural disaster monitoring.
However, traditional centralized processing by the ground server incurs a severe timeliness issue caused by the transmission bottleneck of raw data.
To this end, Space Computing Power Networks (Space-CPN) emerges as a promising architecture to coordinate the computing capability of satellites and enable on-board data processing.
Nevertheless, due to the natural limitations of solar panels, satellite power system is difficult to meet the energy requirements for ever-increasing intelligent computation tasks of artificial neural networks.
To tackle this issue, we propose to employ spiking neural networks (SNNs), which is supported by the neuromorphic computing architecture, for on-board data processing.
The extreme sparsity in its computation enables a high energy efficiency.
Furthermore, to achieve effective training of these on-board models, we put forward a decentralized neuromorphic learning framework, where a communication-efficient inter-plane model aggregation method is developed with the inspiration from RelaySum.
We provide a theoretical analysis to characterize the convergence behavior of the proposed algorithm, which reveals a network diameter related convergence speed.
We then formulate a minimum diameter spanning tree problem on the inter-plane connectivity topology and solve it to further improve the learning performance.
Extensive experiments are conducted to evaluate the superiority of the proposed method over benchmarks.
\end{abstract}

\begin{IEEEkeywords}
Space computing power networks, spiking neural networks, satellite decentralized learning, neuromorphic computing.
\end{IEEEkeywords}

\section{Introduction}
With the advancement of remote sensing technologies, satellites are playing an important role in Earth observation, supporting various downstream tasks.
Typically, these space applications including environment monitoring, disaster warning, and intelligence reconnaissance have strict requirements for response time.
However, conventional approach that direct downloading remote sensing raw data from satellites to the ground for further processing leads to a severe timeliness issue.
This is due to the transmission bottleneck of the relatively low satellite-to-ground link rate (compared with high-speed inter-satellite laser link) and the short visible time window (around 10 minutes) of the access between fast moving satellites and ground stations.
To tackle this challenge, we shall develop Space Computing Power Networks (Space-CPN) by expanding the concept of computing power network~\cite{tang2021computing} with the innovations in satellite networks.
Based on advanced computing payloads and communication techniques on satellites, a space network with the capability of integrated communication and computation can be constructed, where computing power can scheduled flexibly to empower fast on-board data processing.
Nevertheless, a significant challenge faced by the development of Space-CPN lies in the energy bottleneck of satellite power systems, which hinders the implementation of these computation-intensive on-board tasks.
Although satellite power system is experiencing rapid progress, the power supply from solar panels can be easily suspended when satellites move into shaded areas where sunlight cannot cover or encounter eclipse~\cite{alagoz2011energy}.
Even though the energy harvested on the batteries can be leveraged when aforementioned situations happen, frequent usage of batteries inevitably damages the lifespan of satellites. 
Therefore, it is necessary to develop energy-efficient solutions to support on-board data processing in Space-CPN.

Traditionally, satellite on-board computing is implemented on computers with von Neumann architecture.
Such computing architecture is composed of separately deployed memory and processing units.
When performing computation-intensive tasks, data and instructions need to be repeatedly shuttled among these units.
This is the well-known von Neumann bottleneck, and the transmission overheads brought by it inevitably lead to a huge energy cost during data processing.
Comparatively, a recently emerged architecture, neuromorphic computing~\cite{schuman2022opportunities}, leverages the principles of brain processing information. 
Under this architecture, both the capabilities of memory and processing are governed by neurons and synapses, which effectively avoids the overheads brought by von Neumann bottleneck.
Moreover, information is encoded as spikes, whose binary values further enable ultra-low operations.
Therefore, neuromorphic computing acts as a promising architecture to bridge the computation-intensive tasks and low-power environment in Space-CPN.
In addition, the practical implementation of it on satellites has recently been comprehensively evaluated in~\cite{lagunas2024performance}.
Based on this, Spiking Neural Networks (SNNs), which are well adapt to the neuromorphic hardware architecture and can fully unleash the potential of integrated memory and processing, serve as the energy-efficient solution for on-board intelligent models.
The event-driven mechanism and spike information flow of SNNs lead to an extreme sparsity in computation, which further enables energy-efficient on-board data processing in Space-CPN.

To effectively employ these SNNs to serve as intelligent models for on-board data processing, model training is a necessary step.
Conventional ground-based centralized training involves the downloading of massive remote sensing raw data, which leads to a transmission bottleneck and raises severe privacy concern.
To this end, satellite federated learning~\cite{matthiesen2023federated, tao2024federated} becomes a widely adopted approach for distributed model training in Space-CPN.
Specifically, in~\cite{elmahallawy2022asyncfleo, so2022fedspace, lin2024fedsn, wu2024towards}, the authors aim to tackle the model staleness problem to further improve the performance of satellite federated learning.
In~\cite{razmi2022ground, yang2024communication, alsenwi2024flexible}, several methods are proposed to enhance the communication efficiency, while in~\cite{razmi2024energy}, energy consumption acts the primary goal for the system design.
In~\cite{elmahallawy2023optimizing, razmi2022board, razmi2024board}, the authors are devoted to investigate how to effectively leverage inter-satellite links during training.
In~\cite{shi2024satellite}, a comprehensive framework for satellite federated learning is proposed with network flow algorithm used to enhance the transmission between satellites and ground stations.
These methods have shown remarkable performance improvement.
However, they still rely on frequent usage of satellite-to-ground communication links to transmit model information.
The short visible time window issue between satellites and ground stations inevitably leads to a communication bottleneck, which further calls for fully on-board distributed training approaches.

Based on the above observations, to enable low-latency, energy-efficient, and trustworthy on-board data processing in Space-CPN, we propose a brain-inspired decentralized satellite learning framework in this paper.
Specifically, SNNs are adopted to serve as the on-board neuromorphic models for data processing, the high energy efficiency of which effectively break the bottleneck of power consumption for computation-intensive tasks.
Furthermore, to tackle the privacy concern of remote sensing raw data and the communication bottleneck caused by satellite-to-ground links, we consider a fully decentralized learning scheme, where model aggregation is realized via inter-plane inter-satellite links.
Existing inter-plane model aggregation in satellite networks can be generally divided into two categories.
One is global aggregation~\cite{zhai2023fedleo, wu2022dsfl}, which aims to find a inter-plane topology to realize model synchronization on it.
It guarantees the exact global model in each iteration, however, the inter-plane communication overheads directly scales with the size of inter-plane topology.
The other is gossip averaging~\cite{yang2024dfedsat, meng2021decentralized, yan2023convergence, zhou2024decentralized}, where model updates information only exchanged between neighbors of inter-plane topology in each iteration.
This scheme keeps a relatively low inter-plane communication overheads.
Nevertheless, repeatedly averaging makes model updates diffuse slowly across orbit planes, which further degrades the convergence rate.
To tackle the disadvantages of these two schemes, we propose to leverage the idea of RelaySum~\cite{vogels2021relaysum} to achieve inter-plane model aggregation, where model updates are distributed exact uniformly across orbit planes with finite delays depending on the distance between them.
Furthermore, we provide a rigorous theoretical analysis of the convergence behavior of our proposed algorithm, which reveals the relationship between the diameter of the inter-plane topology and the convergence rate.
We then develop a system optimization algorithm to minimize this diameter and improve learning performance.

The major contributions of this paper are summarized as follows:
\begin{itemize}
    \item We propose a novel brain-inspired decentralized satellite learning framework which suits well for Space-CPN.
    SNNs are employed on satellites to enable energy-efficient on-board data processing, and RelaySum is adopted for inter-plane model aggregation to improve communication efficiency and learning performance.
    \item We provide a theoretical convergence analysis of our proposed algorithm framework under non-convex objective, which shows a sublinear convergence rate. Moreover, it is also affected by the diameter of inter-plane topology.
    \item We develop a system optimization algorithm to further improve learning performance. With the minimum diameter serving as the goal, we first model the inter-plane connectivity by analyzing inter-plane inter-satellite links, and then find the optimal spanning tree on it.
    \item We conduct extensive experiments to demonstrate the superiority of the proposed framework compared with existing approaches and the effectiveness of proposed system optimization algorithm.
\end{itemize}
The rest of this paper is organized as follows.
We start by introducing the system model, including satellite constellation overview and decentralized learning procedures, in Section~\ref{system model section}.
Then we present our design of on-board SNNs and RelaySum based inter-plane model aggregation in Section~\ref{algorithm design section}.
Next, we provide theoretical analysis of out proposed framework in Section~\ref{theoretical analysis}.
Subsequently, we provide the system optimization design in Section~\ref{system optimization section}.
Experimental results are then presented in Section~\ref{simulation results section}.
Finally, Section~\ref{conclusion section} concludes this paper.

\begin{figure*}[htbp!]
    \centering
    \includegraphics[width=0.85\linewidth]{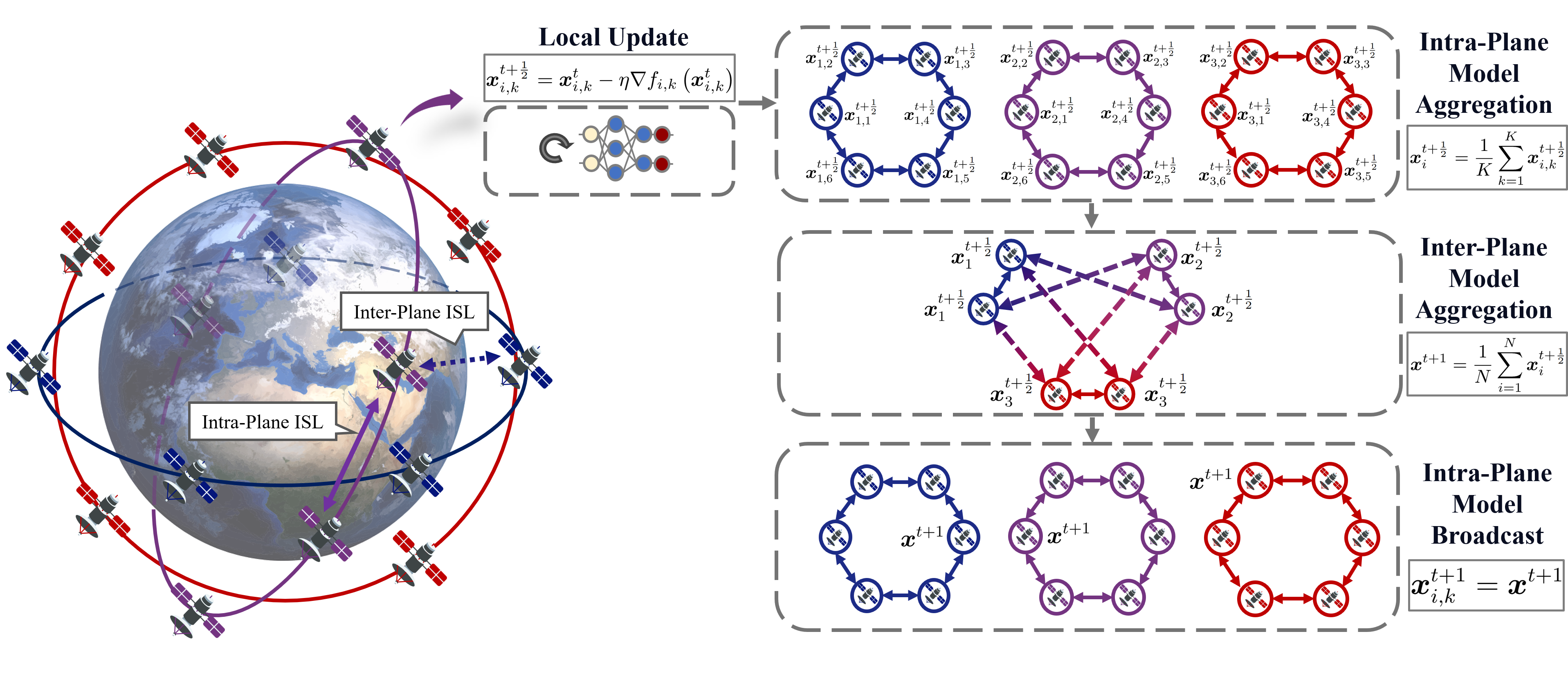}
    \caption{A general decentralized satellite learning system.}
    \label{DecSatFL model}
    \vspace{-1em}
\end{figure*}
\section{System Model}
\label{system model section}
In this section, we first give an overview of satellite networks.
Then we present a general framework of decentralized satellite framework.

\subsection{LEO Constellation}
In this work, we consider a general low-Earth orbit (LEO) constellation to serve as the space network for learning, where satellites are distributed in different circular orbits with various altitudes and inclinations.
Specifically, we use the set $\mathcal{N} := \{1, \cdots i, \cdots N\}$ to denote orbit planes, and each orbit plane $i$ consists of a set of evenly distributed satellites $\mathcal{S}_{i} = \{s_{i,1}, \cdots, s_{i, K}\}$.
The communication between satellites in this space network is implemented via inter-satellite links (ISLs).
They can be generally divided into intra-plane ISLs and inter-plane ISLs.
The former takes place in satellites at the same orbit plane using the antennas located at both sides of the roll axis, and the latter occurs for satellites in different orbit planes using the antennas located at both sides of the pitch axis.
Since the relative position between satellites in the same orbit plane is nearly constant, intra-plane ISLs are usually considered to be stable, while inter-plane ISLs change more frequently due to the diverse movement characteristics of different orbit planes.
Particularly, for two orbits moving in opposite directions, the satellites in these orbits maintain a high relative velocity.
Therefore, the ISLs between such satellites usually suffer from a severe Doppler shift and experience a short visible time window.
This further leads to a poor communication condition and such links are denoted as cross-seam ISLs.

\subsection{Decentralized Satellite Learning}
The goal of this LEO constellation is to complete a learning task based on the raw local data captured by satellites.
Specifically, we use $\mathcal{D}_{i,k}$ to denote the local dataset on each satellite $s_{i,k}$ and the entire dataset $\mathcal{D}$ is represented as the union of all local datasets.
All satellites aim to learn a global model by solving the following empirical risk minimization problem:
\begin{equation}
    \min_{\bm{x} \in \mathbb{R}^{d}} \quad f\bracket{\bm{x}} = \frac{1}{NK}\sum_{i=1}^{N}\sum_{k=1}^{K}f_{i,k}\bracket{\bm{x}},
\end{equation}
where $\bm{x} \in \mathbb{R}^{d}$ is the model parameter, $f\bracket{\cdot}$ acts as the global loss function, $f_{i,k}\bracket{\bm{x}} = \frac{1}{\card{\mathcal{D}_{i,k}}}\ell\bracket{\bm{x};\bm{z}}$ denotes the local empirical loss with $\ell\bracket{\cdot;\bm{z}}$ serving as the loss on sample $\bm{z}$.

To perform this learning task, each satellite $s_{i,k}$ carries out local model update by minimizing $f_{i,k}$, then global model for the next iteration is obtained via averaging these local models.
Conventionally, the aggregation of such global model is enabled by a ground station, with which all satellites upload their local updates.
However, due to the short visible time window issue and resource constraints of ground-to-satellite links (GSLs), the performance of traditional ground-assisted satellite learning is severely limited.
To this end, fully on-board decentralized satellite learning becomes a more promising scheme, where intra-plane and inter-plane ISLs are leveraged to implement global model aggregation and avoid the bottleneck brought by the communication with ground stations.
Specifically, for the $t$-th iteration, $1\leq t \leq T$, the procedures of decentralized satellite learning can be described as follows:
\begin{itemize}
    \item[1)]  \textbf{Local Update}: All satellite perform local model update. 
    In this work, we adopt the common stochastic gradient descent (SGD) method.
    Each satellite $s_{i,k}$ calculates the local gradient based on the local dataset and then updates its model:
    \begin{equation}
        \bm{x}_{i,k}^{t+\frac{1}{2}} = \bm{x}_{i,k}^{t} - \eta \nabla f_{i,k}\bracket{\bm{x}_{i,k}^{t}}, \quad \forall \; i,k,
    \end{equation}
    where $\eta$ denotes the learning rate.
    \item[2)]  \textbf{Intra-Plane Model Aggregation}: Based on the natural ring topology of each orbit plane, the satellites in the same orbit plane performs ring all-reduce~\cite{patarasuk2009bandwidth} to obtain the corresponding intra-plane model updates:
    \begin{equation}
        \bm{x}_{i}^{t+\frac{1}{2}} = \frac{1}{K}\sum_{k=1}^{K}\bm{x}_{i,k}^{t+\frac{1}{2}}, \quad \forall \; i.
    \end{equation}
    \item[3)]  \textbf{Inter-Plane Model Aggregation}: Satellites in different orbit planes share their intra-plane model updates based on inter-plane ISLs and obtain an aggregated global model:
    \begin{equation}
        \bm{x}^{t+1} = \frac{1}{N}\sum_{i=1}^{N} \bm{x}_{i}^{t+\frac{1}{2}}.
    \end{equation}
    \item[4)]  \textbf{Intra-Plane Model Broadcast}: The satellites which have participating the inter-plane model aggregation broadcast the global model across their corresponding orbit planes.
\end{itemize}

The above process exhibits a general framework for decentralized satellite learning, which is shown in Fig.~\ref{DecSatFL model}.
However, it is difficult to directly employ such method in real LEO constellations due to the \textbf{energy constraint} and \textbf{communication bottleneck} faced by the decentralized learning in satellite networks.
We shall elaborate on these issues and propose our corresponding solutions in the next section.

\section{Algorithm Design}
\label{algorithm design section}
In this section, we first give our energy-efficient solution for on-board models.
Next, we provide our communication-efficient approach for inter-plane model aggregation.
Finally, we present the overall algorithm framework for decentralized satellite learning.

\subsection{Brain-Inspired Spiking Neural Network}

As a computation-intensive intelligent model, artificial neural networks (ANNs) bring a huge demand of power consumption.
Even though the power system of satellites is under fast development, this still poses a significant challenge for the effective deployment of ANNs on satellites.
Specifically, since LEOs are all running in the space, there is no outer provision for batteries.
Therefore, the power system of satellites totally depends on equipped solar panels to generate electricity and harvest energy on batteries~\cite{yang2016towards, cheng2022dynamic}.
Particularly, when satellites encounter eclipse or move into shaded areas where sunlight cannot cover, the energy supply from solar panels will suspend~\cite{zhang2023energy, song2022energy}.
On the other hand, even if sufficient harvested energy on batteries is available when aforementioned situation happens, a limit exists on the maximum number of the complete charging cycle for battery cells~\cite{li2024battery}.
Frequent recharging/discharging operations inevitably lead to the acceleration for the exhaustion of satellites' lifespan.
This further reveals the fact of the energy bottleneck on satellites.
In a word, there is a huge gap between the energy supply on satellites and the power consumption of neural networks.

\begin{figure}[htbp!]
    \centering
    \includegraphics[width=0.8\linewidth]{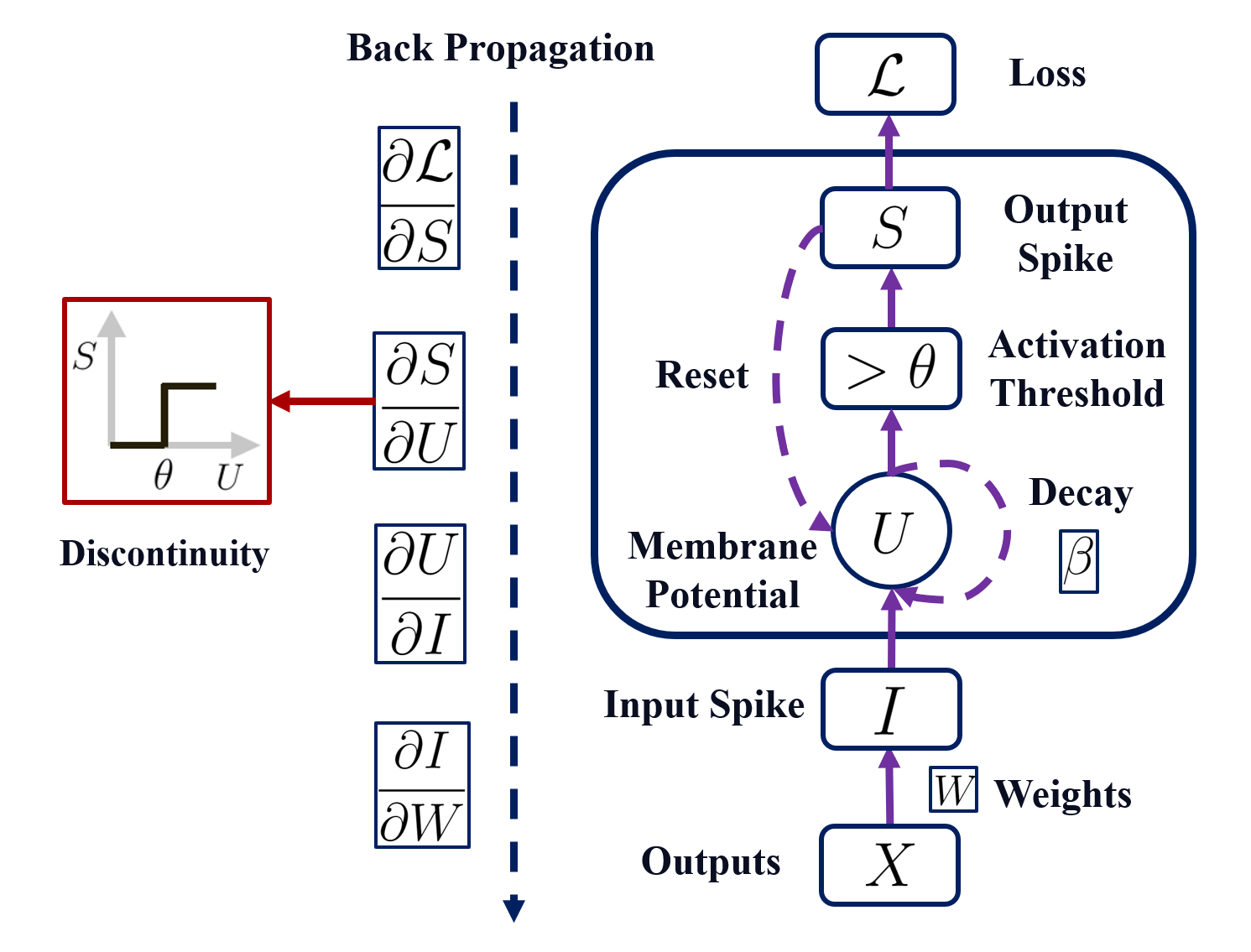}
    \caption{Illustration of the leaky integrate-and-fire (LIF) neuron model.}
    \label{LIF model}
    \vspace{-1em}
\end{figure}

To address this issue, we propose to deploy the third generation neural network, spiking neural networks (SNNs)~\cite{roy2019towards, tavanaei2019deep}, at satellites.
With the strength of the way that brain processing information, it maintains a high energy efficiency and naturally suits the low-power environment in LEO constellations.
To be specific, SNNs share a similar architecture with traditional ANNs, where neurons are interconnected layer by layer with adjustable scalar weights.
However, compared with the static, continuous-valued activation in ANNs, the way neurons in SNNs work is neuromorphic with dynamic and discrete spike activation.
In this work, we consider the most general leaky integrate-and-fire (LIF) neuron model. 
As depicted in Fig.~\ref{LIF model},  such biological neuron leverages membrane potential to track the temporal dynamics of its inner state.
This membrane potential $U[\mathsf{t}]$ is decaying over time and contributed by the weighted sum of the previous layer's output spikes $X[\mathsf{t}]$.
Once the neuron is sufficiently excited, namely its membrane potential reaches a threshold $\theta$, it will emit a spike as output and then reset the membrane potential.
The overall operating principle of each LIF neuron can be characterized as:
\begin{equation}
    U[\mathsf{t}] = \underbrace{\beta U[\mathsf{t-1}]}_{\text{decay}} + \underbrace{WX[\mathsf{t}]}_{\text{input}} - \underbrace{S[\mathsf{t-1}]\theta}_{\text{reset}},
\end{equation}
where $\beta$ and $W$ denote the decaying factor and connection weights, respectively.
$S[\mathsf{t}]$ represents the output spike generated by the neuron, which can be represented as:
\begin{equation}
    S[\mathsf{t}] = \Theta\bracket{U[\mathsf{t}]}= \begin{cases}
        1, \;\;\; U[\mathsf{t}] > \theta,\\
        0, \;\;\; \text{Otherwise}.
    \end{cases}
    \label{output spike with membrane potential}
\end{equation}
It is noteworthy that a majority of neurons stay silence at any time due to this unique biological activation mechanism, and spikes communicated between neurons are all $0-1$ values. 
This means the execution of SNN can be simplified as the transition of discrete and single-bit spikes, which further leads to an extreme sparsity in computation and significantly improves the energy efficiency compared with conventional artificial neural networks.

\begin{figure*}[htbp!]
    \centering
    \includegraphics[width=0.95\linewidth]{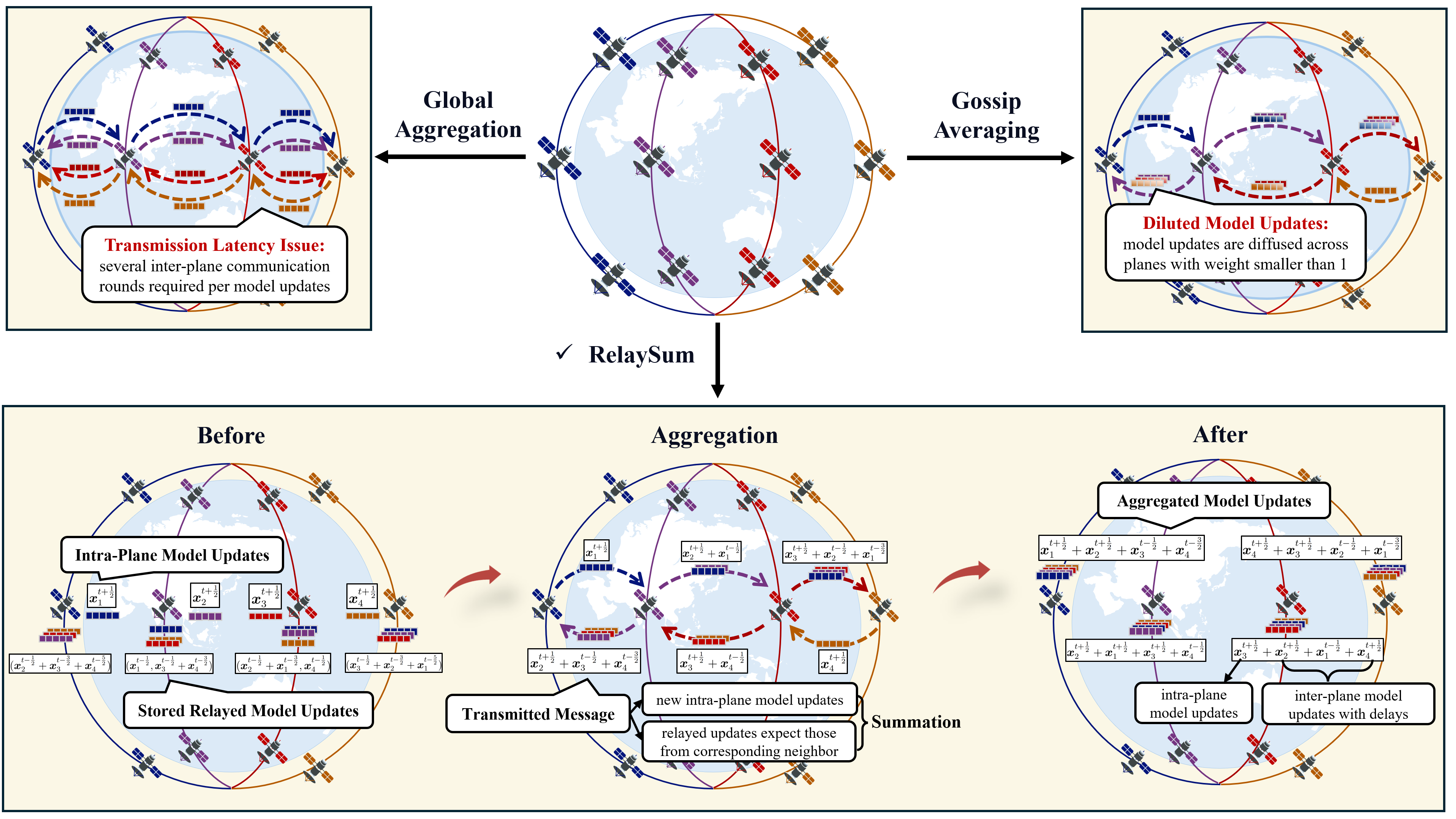}
    \caption{Different inter-plane model aggregation schemes.}
    \label{inter plane model aggregation}
    \vspace{-1em}
\end{figure*}

Though SNNs have shown a great potential in bridging the energy limitation on satellites and the power requirement of neural networks, the effective training for SNNs still poses a serious challenge for its application.
Specifically, there are two mainstream training schemes for SNNs, shadow training and back propagation~\cite{eshraghian2023training}.
The former refers to training a traditional ANN first and then converting it into a SNN, which is easy to implement but suffers from a severe performance loss due to the conversion from high-precision activations to spikes.
The latter follows the classical idea to train neural networks via back propagating the loss and calculating the gradients.
However, new issue arises owing to the unique binary activation mechanism of SNNs.
As illustrated in Fig.~\ref{LIF model}, the output spike $S$ and the membrane potential $U$ actually keep a discontinuous relation, which can be characterized as a Heaviside function in~\eqref{output spike with membrane potential}.
This further leads to a non-differentiable problem during back propagation.
To tackle this issue, most existing solutions leverage the idea of surrogate gradient, where a continuous function is adopted to approximate the original Heaviside function during gradient calculation.
Nevertheless, the introduced smoothing error inevitably degrades the performance in another way.
In order to mitigate such impact, we shall employ a recently proposed method~\cite{wang2023adaptive} to train SNNs.
To be specific, in the training phase we use a coupling of the surrogate function and Heaviside function to serve as a \textbf{hybrid activation} for each LIF neuron:
\begin{equation}
\begin{aligned}
    \tilde{S}[\mathsf{t}] =& (1-\bm{m}) \odot H\bracket{U[\mathsf{t}]} + \bm{m} \odot \Phi\bracket{\Theta\bracket{U[\mathsf{t}]}} \\
                  =& \underbrace{H(U[\mathsf{t}])}_{\text{surrogate}} + \bm{m} \odot \underbrace{\Phi \bracket{\Theta\bracket{U[\mathsf{t}]} - H\bracket{U[\mathsf{t}]}}}_{\text{injected noise without gradient}},
    \label{ASGL function}
\end{aligned}
\end{equation}
where $H(\bm{x})$ and $\Theta(\bm{x})$ denote the surrogate differentiable function and Heaviside activation function, respectively, $\bm{m} \sim \text{Bernoulli}(p)$ represents independent masking with $p$ controlling the probability, $\Phi$ is the function for detaching the gradient and circumventing the non-differentiable problem, namely $\frac{\partial \Phi(\bm{x})}{\partial \bm{x}} = 0$.
This surrogate function $H(\bm{x})$ typically involves trainable parameters to control its shape.
During forward propagation, the effect of the spike activation is characterized as a noise in the second term of~\eqref{ASGL function}, which forces the surrogate fits the Heaviside function.
On the other hand, only the gradient of the surrogate function in the first term of~\eqref{ASGL function} is calculated during back propagation, whose introduced control parameters are leveraged to reduce the smoothing error.
Finally in the inference phase, binary activation is recovered to maintain the spike communication of SNNs.
Based on this special hybrid activation method, the training of SNNs can be efficiently carried out.
Moreover, theoretical guarantee is provided to ensure that optimization via the employment of the hybrid activation~\eqref{ASGL function} will gradually evolve into training a SNN with shared weights.
The core mechanism of this hybrid activation is summarized as in Algorithm~\ref{hybrid activation}.
\begin{algorithm}
    \caption{Core Mechanism of Hybrid Activation}
    \label{hybrid activation}
    \textbf{In Training Phase}: use the following hybrid activation for forward propagation and gradient calculation:
    \begin{align*}
        \tilde{S}[\mathsf{t}] = H(U[\mathsf{t}]) + \bm{m} \odot \Phi \bracket{\Theta\bracket{U[\mathsf{t}]} - H\bracket{U[\mathsf{t}]}}.
    \end{align*}
    
    \textbf{In Inference Phase}: recover the binary activation:
    \begin{align*}
        S[\mathsf{t}] = \Theta\bracket{U[\mathsf{t}]}.
    \end{align*}
\end{algorithm}


\subsection{RelaySum-based Inter-Plane Aggregation}
Based on the neuromorphic learning architecture proposed in the previous subsection, SNN is deployed at satellites and serves as the intelligent model for various downstream tasks.
During local updates, we shall employ the hybrid activation mechanism as in~\eqref{ASGL function} to tackle the non-differentiable problem of SNNs with theoretical performance improvement.
Nevertheless, for the sake of a generalized and high-performance intelligent model, the cooperation between satellites is a necessary step in decentralized federated learning and it can be considered from two aspects as illustrated in Section II-B.
As for the intra-plane model aggregation, the satellites in the same orbit move towards the same direction with fixed velocities.
This leads a relatively stationary position with stable channel characteristics between adjacent satellites. 
Therefore, the satellites in the same orbit plane naturally form a ring communication topology with reliable transmission, which makes it reasonable to use ring all-reduce algorithm to achieve efficient intra-plane model aggregation.
On the other hand, for the inter-plane model aggregation, due to the diverse movement characteristics of different orbit planes (e.g., direction, inclination, and altitude), inter-plane ISLs are more unstable compared with intra-plane ISLs and their link performance is severely affected by the Doppler shift.
This results in the difficulty to directly collect the model updates across different orbit planes via inter-plane ISLs.
Therefore, how to design an effective inter-plane model aggregation scheme becomes the key to decentralized satellite federated learning.

Existing works of decentralized satellite federated learning can be divided into two categories with respect to how model information spreads across different planes.
The former aims at global aggregation~\cite{zhai2023fedleo, wu2022dsfl}, where a global model is aggregated across all orbits in each iteration.
It is equivalent to leverage ground stations to perform centralized aggregation and also suffers from the same problem of communication bottleneck.
The existence of cross-seam ISLs with poor link performance directly hinders the formulation of nice topology like ring where efficient aggregation methods can be employed, and using ground stations to serve as relay for such poor links would be affected by the inefficiency of GSLs.
Therefore, only a general inter-plane tree topology can be formulated and the global model aggregation on it incurs a significant latency issue, which scales with the size of the constellation.
The latter targets on partial aggregation~\cite{yang2024dfedsat, meng2021decentralized, yan2023convergence, zhou2024decentralized}, where all orbit planes only interact with their neighbors on the formulated communication topology in each iteration.
Based on the principle of gossip averaging, the communication overheads can be largely reduced compared with the previous scheme.
Nevertheless, the model updates diffuse slowly across orbit planes due to repeated averaging with weights smaller than $1$, which further leads to a degradation on convergence and the sensitiveness for data heterogeneity of different orbits.

To tackle the drawbacks of the above two approaches, we adopt the idea of RelaySum~\cite{vogels2021relaysum} to realize inter-plane model aggregation.
As illustrated in Fig.~\ref{inter plane model aggregation}, RelaySum maintains the same communication scheme with gossip averaging, where model updates are only exchanged between neighbor orbits so that the communication overheads can be kept at a relatively low level.
However, as we have discussed before, for two far apart orbits, the model updates can be exponentially weakened with gossip averaging, which further slows down the convergence rate and leads to a higher total communication overheads.
To avoid such issue, instead of iteratively averaging model updates, RelaySum let each orbit relay messages of model updates on the inter-plane topology without decaying.
More Specifically, through the usage of additional memory, each orbit stores the summation of received model updates and transmits tailored messages (the summation of new intra-plane updates and stored updates except that from target neighbor) to its neighbors.
In this way, in each iteration, all orbits can obtain a uniform average of exactly one model updates from each orbit with the delay corresponding to the distance on the inter-plane routing tree:
\begin{equation}
    \bm{x}_{i}^{t} = \frac{1}{N}\sum_{j=1}^{N}\bm{x}_{j}^{t-\tau_{ij}+\frac{1}{2}},
\end{equation}
where $\bm{x}_{i}^{t+\frac{1}{2}}$ denotes the intra-plane model updates of orbit $i$ in the $t$-iteration, $\tau_{ij}$ measures the distance of two orbits on the inter-plane routing tree.
Specifically, we have $\tau_{ii}=0$ and $\tau_{ij}$ equals to the minimum network hops from orbit $i$ to $j$ minus $1$. 
Based on this RelaySum inter-plane aggregation approach, a better convergence behavior of training can be achieved compared with classical gossip averaging.
On the other hand, it suits for any decentralized networks so that inter-plane ISLs with poor performance such as cross-seam ISLs can be avoided in the construction of inter-plane aggregation topology.

In summary, based on the spiking neural network deployed at satellite-side with the corresponding training mechanism introduced in Section III-A, and the inter-plane model cooperation scheme employed in Section III-B, we propose our brain-inspired decentralized satellite learning framework as in Algorithm~\ref{DecSatSNN}.
\begin{algorithm}
    \caption{Proposed Brain-Inspired Decentralized Satellite Learning Algorithm}
    \label{DecSatSNN}
    \textbf{Initialization}: inter-plane routing tree $\mathcal{T}$; learning rate $\eta$; initial models: $\bm{x}_{i}^{0} = \overline{\bm{x}}^{0}, \forall i\in\mathcal{N}$; auxiliary variables: $\bm{b}_{i,j}^{-1}=\bm{0},\; c_{i,j}^{-1}=0, \;\; \forall\;i,j\in\mathcal{N}$.\\
    \Foreach{global iteration $t = 0, 1, \cdots, T-1$}{
        \Forallpara{orbit $i \in \mathcal{N}$}{
            \Forallpara{satellite $s_{i,k} \in \mathcal{K}_{i}$}{
                $\bm{x}_{i,k}^{t,0,0} \leftarrow \bm{x}_{i}^{t}$;
            }
        }
        \Foreach{intra-orbit round $r = 0, 1, \cdots, R-1$}{
            \Forallpara{orbit $i \in \mathcal{N}$}{
                \Forallpara{satellite $s_{i,k} \in \mathcal{K}_{i}$}{
                    \Foreach{local epoch $e = 0, 1, \cdots, E-1$}{
                        $\bm{x}_{i,k}^{t,r,e+1} \leftarrow \bm{x}_{i,k}^{t,r,e} - \eta\nabla F_{i,k}\bracket{\bm{x}_{i,k}^{t,r,e}}$;
                    }
                }
                 use ring all-reduce algorithm to calculate the intra-orbit model: $\bm{x}_{i}^{t,r+1} \leftarrow \frac{1}{K}\sum_{k=1}^{K}\bm{x}_{i,k}^{t,r,E-1}$;\\
                \Forallpara{satellite $s_{i,k} \in \mathcal{K}_{i}$}{
                    $\bm{x}_{i,k}^{t,r+1,0} \leftarrow \bm{x}_{i}^{t,r+1}$;
                }
            }
           
        }
        \Forallpara{orbit $i \in \mathcal{N}$}{
            $\bm{x}_{i}^{t+\frac{1}{2}} \leftarrow \bm{x}_{i}^{t,R-1}$; \\
            \Foreach{neighbor $j \in \mathcal{N}_{i}$}{
                transmit relayed updates: $\bm{b}_{i,j}^{t} = \bm{x}_{i}^{t+\frac{1}{2}}+\sum_{l\in\mathcal{N}_{i}\backslash j}\bm{b}_{l,i}^{t-1}$, \\
                and relayed counters: $c_{i,j}^{t} = 1 + \sum_{l\in\mathcal{N}_{i}\backslash j}c_{l,i}^{t-1}$ to orbit $j$;\\
                receive relayed updates and counters $\bracket{\bm{b}_{j,i}^{t}, c_{j,i}^{t}}$ from orbit $j$;
            }
            $n_{i}^{t} \leftarrow 1 + \sum_{j\in\mathcal{N}_{i}}c_{j,i}^{t}$;\\
            $\bm{x}_{i}^{t+1} \leftarrow \frac{1}{n_{i}^{t}}\bracket{\bm{x}_{i}^{t+\frac{1}{2}}+\sum_{j\in\mathcal{N}_{i}}\bm{b}_{j,i}^{t}}$;
        }
    }
\end{algorithm}

\section{Theoretical Analysis}
\label{theoretical analysis}
In this section, we shall provide the theoretical guarantee of the proposed brain-inspired decentralized satellite learning algorithm.
To begin with, we first present some preliminaries for the following analysis.

\subsection{Preliminaries}
First of all, since we are using the hybrid activation~\eqref{ASGL function} in the training phase, the relationship between the prototype noisy neural network with such activation and the actual SNN with binary activation serves as the cornerstone of our theoretical analysis.
Here, we invoke the theorem provided in~\cite{wang2023adaptive}:
\begin{theorem}
    Minimizing the loss of the noisy neural network $f^{n}$ can be approximated into minimizing the loss of the embedded SNN $f^{s}$, regularized by the layerwise distance between the surrogate $H(\cdot)$ and the Heaviside function $\Theta(\cdot)$:
    \begin{equation}
    \begin{aligned}
        &f^{n}\bracket{\bm{x}} \approx f^{s}\bracket{\bm{x}} + \frac{1-p}{2p}\sum_{l=1}^{L}\langle D^{l}, \text{diag}\bracket{H\bracket{U^{l}}-\Theta\bracket{U^{l}}}^{\odot2} \rangle,
    \end{aligned}
    \end{equation}
    where $L$ denotes the total layer of the neural network, $D^{l}$ is a constant related to the second derivative of the $l$-layer, $U^{l}$ represents the integrated membrane potential of the $l$-th layer.
\end{theorem}
It is observed that this theorem essentially provides a theoretical guarantee that the training over the noisy neural network with hybrid activation ensures the minimization for the loss of the embedded SNN, as well as the reduction for the smoothing error of the surrogate function.
Therefore, in the following analysis we shall focus on the loss of the prototype neural network $f^{n}$, and we will omit the superscript for convenience.

Based the proposed algorithm framework, we first have these properties:
\begin{itemize}
    \item[1)] The local loss of each satellite $f_{i,k}(\cdot)$, the intra-orbit loss of each orbit $f_{i}(\cdot)$, and the global loss $f(\cdot)$ have following relationships:
    \begin{equation}
        f_{i}\bracket{\bm{x}} = \frac{1}{K}\sum_{k=1}^{K}f_{i,k}\bracket{\bm{x}}, \;\; f\bracket{\bm{x}} = \frac{1}{N}\sum_{i=1}^{N}f_{i}\bracket{\bm{x}}.
    \end{equation}
    \item[2)] The intra-orbit updates can be written as:
    \begin{equation}
        \bm{x}_{i}^{t+\frac{1}{2}} = \bm{x}_{i}^{t} - \frac{\eta}{K}\sum_{r=0}^{R-1}\sum_{k=1}^{K}\sum_{e=0}^{E-1}\nabla F_{i,k}\bracket{\bm{x}_{i,k}^{t,r,e}}.
    \end{equation}
\end{itemize}

\noindent\textbf{Assumption 1} ($L$-Smoothness): The local loss of each satellite $f_{i,k}(\cdot)$, the intra-orbit loss of each orbit $f_{i}(\cdot)$, and the global loss $f(\cdot)$ are all $L$-smooth.

\noindent\textbf{Assumption 2} (Unbiased Gradient with Bounded Variance): Each satellite can calculate an unbiased stochastic gradient with bounded variance, namely, for all $i\in\mathcal{N}, k\in\mathcal{K}_{i}$:
$$
\mathbb{E}\mbracket{\nabla F_{i,k}\bracket{\bm{x}}} = \nabla f_{i,k}\bracket{\bm{x}}, 
$$
$$
\mathbb{E}\mbracket{\norm{\nabla F_{i,k}\bracket{\bm{x}}}^{2} - \norm{\nabla f_{i,k}\bracket{\bm{x}}}^{2}} \leq \sigma^{2}.
$$

\noindent\textbf{Assumption 3} (Intra-Orbit Dissimilarity): For each orbit $i \in \mathcal{N}$, there exists a constant $\delta$ such that:
$$
\frac{1}{K}\sum_{k=1}^{K}\norm{\nabla f_{i,k}\bracket{\bm{x}}- \nabla f_{i}\bracket{\bm{x}}}^{2} \leq \delta^{2}.
$$

\noindent\textbf{Assumption 4} (Inter-Orbit Dissimilarity): For each orbit $i \in \mathcal{N}$, there exists a constant $\zeta$ such that:
$$
\norm{\nabla f_{i}\bracket{\bm{x}} - \nabla f\bracket{\bm{x}}}^{2} \leq \zeta^{2}.
$$

We use $\tau_{ij}$ to denote the number of hops between orbit $i$ and $j$ minus $1$ in the given inter-plane routing tree $\mathcal{T}$, and we have $\tau_{max} = \max_{ij}\tau_{ij}$.
Besides, we use $t_{ij}$ to denote $t-\tau_{ij}$.

To track the model and gradient information across different orbits, we define the following matrices:
\begin{equation}
    \bm{X}^{t} = \mbracket{\bm{x}_{1}^{t}, \bm{x}_{2}^{t}, \cdots, \bm{x}_{N}^{t}}^{\transpose} \in \mathbb{R}^{N\times d},
\end{equation}
\begin{equation}
    \nabla\bm{F}^{t} = \mbracket{\nabla F_{1}^{t}, \nabla F_{2}^{t}, \cdots, \nabla F_{N}^{t}}^{\transpose} \in \mathbb{R}^{N\times d},
\end{equation}
where $\nabla F_{i}^{t} = \frac{1}{K}\sum_{r=0}^{R-1}\sum_{k=1}^{K}\sum_{e=0}^{E-1}\nabla F_{i,k}\bracket{\bm{x}_{i,k}^{t,r,e}}$ represents the effective gradient for orbit $i$ in $t$-th iteration.

Furthermore, to characterize the delay impact on the relayed information, we define the following stacked model and gradient information:
\begin{align}
    \bm{Y}^{t} = \mbracket{\begin{array}{c}
         \bm{X}^{t} \\
         \bm{X}^{t-1} \\
         \cdots \\
         \bm{X}^{t-\tau_{max}}
    \end{array}}, 
    \bm{G}^{t} = \mbracket{\begin{array}{c}
         \nabla\bm{F}^{t} \\
         \nabla\bm{F}^{t-1} \\
         \cdots \\
         \nabla\bm{F}^{t-\tau_{max}}
    \end{array}} \in \mathbb{R}^{N(\tau_{max}+1)\times d}
\end{align}

Then the model updates of the proposed framework can be alternatively written as
\begin{equation}
    \bm{Y}^{t+1} = \bm{W}\bm{Y}^{t} - \eta\tilde{\bm{W}}\bm{G}^{t},
    \label{simplified model updates}
\end{equation}
where $\bm{W}, \tilde{\bm{W}} \in \mathbb{R}^{N(\tau_{max}+1)\times N(\tau_{max}+1)}$ serve as the mixing matrices for model and gradient information, which are defined as follows:
\begin{equation}
    \mbracket{\bm{W}}_{N\tau+i,N\tau^{\prime}+j} = \begin{cases}
        \frac{1}{N}\;\; \tau=0 \;\text{and}\; \tau^{\prime}=\tau_{ij}\\
        1\;\; i=j \;\text{and}\; \tau = \tau^{\prime} + 1 \\
        0 \;\; \text{otherwise}
    \end{cases}
\end{equation}
\begin{equation}
    \mbracket{\tilde{\bm{W}}}_{N\tau+i,N\tau^{\prime}+j} = \begin{cases}
        \frac{1}{N} \;\; \tau=0 \;\text{and}\; \tau^{\prime}=\tau_{ij}\\
        0 \;\; \text{otherwise}
    \end{cases}
\end{equation}
Particularly, the model mixing matrix $\bm{W}$ is row stochastic. 
The right eigenvector of $\bm{W}$ is $\bm{1}_{N(\tau_{max}+1)}$ and we use $\bm{\pi} = [\bm{\pi}_{0}^{\transpose}, \cdots, \bm{\pi}_{\tau_{max}}^{\transpose}]^{\transpose}\in \mathbb{R}^{N(\tau_{max}+1)}$ to denote its left eigenvector such that $\bm{\pi}^{\transpose}\bm{1}_{N(\tau_{max}+1)} = 1$, where $\bm{\pi}_{0}$ is a vector of same value $\pi_{0}$ according to~\cite[Lemma 4]{vogels2021relaysum}.

To characterize the convergence behavior of the proposed framework, we consider the following metrics:
\begin{itemize}
    \item[1)] expected function suboptimality: $r_{t} := \mathbb{E}f\bracket{\overline{\bm{x}}^{t}} - f^{*}$;
    \item[2)] consensus distance: $\theta_{t} := \frac{1}{N}\Fnorm{\overline{\bm{Y}}^{t} - \bm{Y}^{t}}^{2}$;
    \item[3)] gradient squared norm: $e_{t} := \norm{\nabla f\bracket{\overline{\bm{x}}^{t}}}^{2}$,
\end{itemize}
where $\overline{\bm{x}}^{t} = \bracket{\bm{\pi}^{\transpose}\bm{Y}^{t}}^{\transpose}$ denotes the average model, $f^{*}$ be the optimal objective value, and $\overline{\bm{Y}}^{t} = \bm{1}\bm{\pi}^{\transpose}\bm{Y}^{t}$.

\subsection{Convergence Analysis}

Since we are using stochastic gradient descent to perform local update, the noise introduced in this step shall have an impact on the convergence behavior.
Therefore, we first provide a lemma to track the effect caused by such noise. 
\begin{lemma}
    (\emph{Bound of Stochastic Noise}): Let Assumption 2 hold, then we have the following relationships between stochastic gradient and true gradient.
    \begin{equation}
        \mathbb{E}\norm{\bm{\pi}^{\transpose}\tilde{\bm{W}}\bracket{\bm{G}^{t}-\mathbb{E}\bm{G}^{t}}}^{2} \leq NR^{2}E^{2}\pi_{0}^{2}\sigma^{2},
    \end{equation}
    \begin{equation}
        \mathbb{E}\norm{\tilde{\bm{W}}\bracket{\bm{G}^{t}-\mathbb{E}\bm{G}^{t}}}^{2} \leq R^{2}E^{2}\sigma^{2}.
    \end{equation}
\end{lemma}

The proof can be found in Appendix~\ref{lemma 1 proof}. 
Then we proceed on presenting the dynamics of the first performance metric, expected function suboptimality.

\begin{lemma}
    (\emph{Decrease of the Non-Convex Objective}): Let Assumption 1-4 hold, and the learning rate $\eta \leq \frac{\tilde{\pi}_{0}}{36REL}$ with $\tilde{\pi}_{0} = \min\{\pi_{0}, 1\}$, we have the following descent property for the first metric which characterizes the expected function suboptimality:
    \begin{equation}
    \begin{aligned}
        r_{t+1} \leq &  r_{t} - \frac{NRE\pi_{0}\eta}{4}e_{t}+2NRE\pi_{0}\eta L^{2}\theta_{t}\\
        &+NR^{2}E^{2}\pi_{0}^{2}\eta^{2}L\sigma^{2}+5\bracket{D_{1}+D_{2}}NRE\pi_{0}\eta z^{2},
        \label{bound of non-convex objective}
    \end{aligned}
    \end{equation}
    where
    \begin{equation}
        D_{1} = 7\eta^{2}E(E-1)L^{2}, \;\; D_{2} = 7\eta^{2}E^{2}R(R-1)L^{2},        
    \end{equation}
    \begin{equation}
        z^{2} = \sigma^{2}+\delta^{2}+\zeta^{2}.
    \end{equation}
\end{lemma}

The proof can be found in Appendix~\ref{lemma 2 proof}.
Next, we shall provide the analysis of the consensus distance, and we first invoke an important lemma as follows. 

\begin{lemma}
    (\emph{Property of Mixing Matrix, Lemma 1 in~\cite{vogels2021relaysum}}). There exists an integer $m=m(\bm{W}) > 0$ such that for any $\bm{X} \in \mathbb{R}^{N(\tau_{max}+1)\times d}$ we have
    \begin{equation}
        \norm{\bm{W}^{m}\bm{X}-\bm{1}\bm{\pi}^{\transpose}\bm{X}}^{2} \leq \bracket{1-q}^{2m}\norm{\bm{X}-\bm{1}\bm{\pi}^{\transpose}\bm{X}}^{2},
    \end{equation}
    where $q = \frac{1}{2}(1-|\lambda_{2}(\bm{W})|)$, and we use $\rho := \frac{q}{m}$ to denote the effective spectral gap of $\bm{W}$.
\end{lemma}

Then we formally give our result for the dynamics of consensus distance during training:
\begin{lemma}
    (\emph{Bound of the Consensus Distance}): Let Assumption 1-4 hold, we have the following property for the second metric which characterize the consensus distance:
    \begin{equation}
    \begin{aligned}
        \frac{1}{T}\sum_{t=0}^{T-1}\theta_{t} \leq& \frac{20C_{1}^{2}m^{2}R^{2}E^{2}\eta^{2}}{q^{2}}\frac{1}{T}\sum_{t=0}^{T-1}e_{t}+\frac{8C_{1}^{2}m^{2}R^{2}E^{2}\eta^{2}}{Nq^{2}}\sigma^{2}\\
        &+\frac{144(D_{1}+D_{2})C_{1}^{2}m^{2}R^{2}E^{2}\eta^{2}}{q^{2}}z^{2},
        \label{bound of consensus distance}
    \end{aligned}
    \end{equation}
    where $C_{1}$ is a constant defined as~\cite[Definition G]{vogels2021relaysum}:
    \begin{equation}
    \begin{aligned}
        C_{1}^{2} :=& \max_{i=0,\cdots,m-1}\norm{\bm{W}^{i}\tilde{\bm{I}}}^{2} = C^{2}\norm{\bm{1}\bm{\pi}^{\transpose}\tilde{\bm{I}}}^{2} \\
        =& C^{2}N^{2}(\tau_{max}+1)\pi_{0}^{2},
        \label{definition C}
    \end{aligned}
    \end{equation}
    with $\tilde{\bm{I}} \in \mathbb{R}^{N(\tau_{max}+1)\times N(\tau_{max}+1)}$ satisfies
    \begin{align*}
        [\tilde{\bm{I}}]_{ij} = \begin{cases}
            1 \quad i=j\leq N,\\
            0 \quad \text{otherwise}.
        \end{cases}
    \end{align*}
\end{lemma}

The proof can be found in Appendix~\ref{lemma 4 proof}.
Based on the theoretical results obtained with the first two metrics, we can conclude the final convergence guarantee with respect to the third metric for the proposed framework.
\begin{theorem}
    (\emph{Convergence Guarantee in the Non-Convex Case}). Let Assumption 1-4 hold, for a learning rate $\eta < \frac{q\tilde{\pi}_{0}}{36C_{1}mREL}$ with $\tilde{\pi}_{0} = \min\{\pi_{0}, 1\}$, $\tilde{\tau} = \tau_{max}+1$, we have:
    \begin{equation}
    \label{convergence behavior 1}
    \begin{aligned}
        \frac{1}{T}\sum_{t=0}^{T-1}\norm{\nabla f\bracket{\overline{\bm{x}}^{t}}}^{2} \leq& 16\bracket{\frac{2L\sigma^{2}r_{0}}{NT}}^{\frac{1}{2}}+16\bracket{\frac{4C\sqrt{\tilde{\tau}}L\sigma r_{0}}{\rho\sqrt{N}T}}^{\frac{2}{3}}\\
        &+\frac{288CL\sqrt{\tilde{\tau}}r_{0}}{\rho T} \\
        &+16\Bigg[\frac{\sqrt{7E(E-1)+7E^{2}R(R-1)}}{NRE\pi_{0}}\\
        &\cdot\sqrt{\frac{2C^{2}\tilde{\tau}}{9\rho^{2}L^{2}}+5}\frac{Lzr_{0}}{T}\Bigg]^{\frac{2}{3}},
    \end{aligned}
    \end{equation}
    where constant $C$ is given in~\eqref{definition C} and $r_{0} = f(\overline{\bm{x}}^{0}) - f^{*}$ denotes the optimality gap of the initial point.
    Alternatively, it needs 
    \begin{equation}
        \label{convergence behavior 2}
        \mathcal{O}\bracket{\frac{\sigma^{2}}{N\epsilon^{2}}+\frac{C\sqrt{\tilde{\tau}}\sigma}{\rho\sqrt{N}\epsilon^{\frac{3}{2}}}+\frac{C\sqrt{\tilde{\tau}}}{\rho\epsilon}+\frac{C\sqrt{\tilde{\tau}\bracket{\sigma^{2}+\delta^{2}+\zeta^{2}}}}{\rho\sqrt{N} L\epsilon^{\frac{3}{2}}}}Lr_{0}
    \end{equation}
    iterations to reach any given target accuracy $\epsilon$.
\end{theorem}

The proof can be found in Appendix~\ref{theorem 1 proof}.

\noindent\textbf{Remark 1.}
\emph{It can be observed that the proposed algorithm maintains a sublinear convergence rate, and by setting $E=1$ and $R=1$, the convergence behavior degenerates to a similar case with the original RelaySum~\cite{vogels2021relaysum} as in~\eqref{convergence behavior 1}.
Moreover, as illustrated in~\eqref{convergence behavior 2}, $\mathcal{O}\bracket{\frac{\sigma^{2}}{N\epsilon^{2}}}$ serves as the dominant term and is consistent with centralized SGD.
On the other hand, the rest three terms are all affected by $\tilde{\tau} = \tau_{max}+1$, which essentially characterizes the diameter of the topology for inter-plane model aggregation.
This further convinces an intuitive fact that the longer diameter of the routing tree is, the more number of iterations is required to diffuse the model updates across the network.}

\section{System Optimization}
\label{system optimization section}
In this section, we focus on the system optimization of our proposed algorithm to further improve the learning performance.

\subsection{Problem Formulation}
According to the theoretical result given in the previous section, the diameter of the inter-plane routing tree employed for model aggregation, namely $\tilde{\tau}$, plays a crucial role on the convergence behavior of the proposed framework.
Therefore, for the sake of a faster convergence rate, the diameter of inter-plane routing tree should be as short as possible with the coverage of all orbits.
To find such routing tree for model aggregation, we first characterize the connectivity of orbit planes.

As we have discussed before, since the movement characteristics vary among different orbit planes, inter-plane ISLs are much more unstable compared with intra-plane ISLs and exhibit a dynamic evolution over time.
Typically, a feasible inter-plane ISL are mainly determined by two factors, line-of-sight (LoS) distance and Doppler shift effect~\cite{leyva2021inter}.
The former one characterizes the static relation between two satellites.
Given two satellites $u$ and $v$ with their position information being $\bracket{\varsigma_{u}, \vartheta_{u}}$ and $\bracket{\varsigma_{v}, \vartheta_{v}}$, respectively, where $\varsigma$ denotes latitude and $\vartheta$ represents longitude.
The geocentric angle between these two satellites can be written as~\cite{zhu2022laser, chen2024shortest}:
\begin{equation}
    \varphi = \arccos\mbracket{\sin\bracket{\varsigma_{u}}\sin\bracket{\varsigma_{v}}+\cos\bracket{\varsigma_{u}}\cos\bracket{\varsigma_{v}}\cos\bracket{\vartheta_{u}-\vartheta_{v}}}.
\end{equation}
Then assume the Earth is a perfect sphere with $r_{E}$ being its radius, and let $h_{u}$ and $h_{v}$ be the orbit plane's altitudes of these two satellites, the distance between them can be calculated as:
\begin{equation}
    d_{u,v} = \sqrt{\tilde{h}_{u}^{2}+\tilde{h}_{v}^{2}-2\tilde{h}_{u}\tilde{h}_{v}\cos\bracket{\varphi}},
\end{equation}
where $\tilde{h}_{u} = h_{u}+r_{E}$ and $\tilde{h}_{v} = h_{v}+r_{E}$.
On the other hand, the maximum slant range between satellite $u$ and $v$ can be given as:
\begin{equation}
    d_{u,v}^{*} = \sqrt{h_{u}\bracket{h_{u}+2r_{E}}}+\sqrt{h_{v}\bracket{h_{v}+2r_{E}}}.
\end{equation}
Therefore, for two satellites $u$ and $v$ whose LoS is not blocked the Earth, namely, are visible to each other, their interplanetary distance must satisfy the following condition:
\begin{equation}
    d_{u,v} \leq d_{u,v}^{*}.
\end{equation}
The latter factor, Doppler shift, characterizes the dynamic relation between satellites, and it can be formally given as~\cite{leyva2020leo, boumalek2024leo, liu2013doppler}:
\begin{equation}
    f_{u,v} = \psi_{u,v} f_{c} / c, 
\end{equation}
where $\psi_{u,v}$ denotes the relative speed between satellite $u$ and $v$, $f_{c}$ represents the carrier frequency, and $c$ is the speed of light. 
Such Doppler shift $f_{u,v}$ leads to a carrier frequency offset between satellites, which inevitably degrades the communication performance.
Typically, various compensation techniques can be employed to mitigate this effect.
Nevertheless, for two satellites experiencing a severe Doppler shift, it is quite difficult to solve this issue and reliable ISL between them are usually considered to be infeasible~\cite{leyva2021inter, pi2022dynamic}.
Here, we use constant $f_{max}$ to denote the maximum tolerable value for Doppler shift.

Combine the above analysis of the two factors, we can formally give the following definition.
\begin{definition}
    (Eligible Pair). For a given timestamp, an eligible pair $\bracket{u,v}$ represents the existence of reliable inter-satellite link between satellite $u$ and $v$, which satisfies $d_{u,v} \leq d_{u,v}^{*}$ and $f_{u,v} \leq f_{max}$.
\end{definition}
Then we can characterize the connectivity among orbit planes as follows:
\begin{definition}
    (Inter-Plane Connectivity). The connectivity among different orbit planes can be modelled as an undirected graph $\mathcal{G} = \bracket{\mathcal{V}, \mathcal{E}}$, where $\mathcal{V} = \mathcal{N}$ is the set of orbit planes serving as vertices in the graph and $\mathcal{E}$ denotes the set of edges between orbit planes.
    Particularly, $e_{i,j} \in \mathcal{E}$ if and only if there exists an eligible pair between the satellites of orbit $i$ and $j$ for any timestamp.
\end{definition}

This actually provides a stable inter-plane communication topology on the given constellation, and we need to construct a routing tree on it to serve as the basis for inter-plane model aggregation.
Take the minimum diameter $\tilde{\tau}$ as the goal for a better learning performance, we can formulate the following problem:
\begin{equation}
    \mathscr{P}_{1}: \quad\quad \underset{\mathcal{T}=\bracket{\mathcal{V}, \tilde{\mathcal{E}}}}{\text{minimize}}  \quad \tilde{\tau},
\end{equation}
where $\mathcal{T}$ is a spanning tree over $\mathcal{G}$ with $\tilde{\mathcal{E}} \subseteq \mathcal{E}$.

\subsection{Routing Tree Optimization}

It can be observed that $\mathscr{P}_{1}$ is a \emph{minimum diameter spanning tree} (MDST) problem in essence, and we can directly seek the optimal spanning tree $\mathcal{T}$ based on $\mathscr{P}_{1}$.
Nevertheless, there may exist multiple spanning trees maintain the same minimum diameter, which means simply relying on the diameter factor is not able to determine the optimal one in this case.
To tackle this issue, we propose to leverage the quality of communication links between different orbit planes to serve as an extra impact factor.

To be specific, for a given eligible pair $\bracket{u, v}$, the free space path loss between $u$ and $v$ can be modelled as:
\begin{equation}
    L_{u,v} = \bracket{\frac{4\pi d_{u,v}f_{c}}{c}}^{2},
\end{equation}
The performance of the ISL between $u$ and $v$ can then be characterized by the following expression of SNR~\cite{leyva2021inter}:
\begin{equation}
    \text{SNR}_{u,v} = \frac{\text{EIRPG}}{\kappa \varrho B L_{u,v}},
\end{equation}
where EIRPG is the effective isotropic radiated power plus receiver antenna gain, $\kappa$ is Boltzmann constant, $\varrho$ is the thermal noise, and $B$ denotes the bandwidth.
Then the communication quality between two orbits can be characterized by the average SNR of the eligible pairs:
\begin{equation}
    \xi_{i,j} = \frac{1}{|\mathcal{I}_{i,j}|}\sum_{(u,v)\in \mathcal{I}_{i,j}}\text{SNR}_{u,v},
    \label{communication quality}
\end{equation}
where $\mathcal{I}_{i,j}$ denotes the set of eligible pairs between plane $i$ and $j$.
It is noteworthy that the quality of inter-plane communication link $\xi_{i,j}$ only serves as a second impact factor which affects the communication in a single aggregation round, while the diameter of the inter-plane routing tree $\tilde{\tau}$ is still the dominant term that affects the overall convergence rate.
Based on this principle, we assign the weights for edges on the inter-plane connectivity graph $\mathcal{G}$ as follows:
\begin{equation}
    w\bracket{e_{i,j}} = \tilde{\xi}+\frac{1}{\xi_{i,j}}, 
    \label{weights of inter-plane topology}
\end{equation}
where $\tilde{\xi} = \sum_{e_{i^{\prime},j^{\prime}}\in\mathcal{E}}\frac{1}{\xi_{i^{\prime},j^{\prime}}}$ denotes the summation of the inversion of communication quality factors, and we have the following proposition:

\emph{Proposition 1}. For any two spanning trees $\mathcal{T}_{1}$ and $\mathcal{T}_{2}$ over graph $\mathcal{G}$, if the diameter of $\mathcal{T}_{1}$ is smaller than $\mathcal{T}_{2}$ in the case of all edges' weights set to be 1, then same result holds in the case of all edges' weights set as in~\eqref{weights of inter-plane topology}.

\emph{Proof}. Suppose the diameters of $\mathcal{T}_{1}$ and $\mathcal{T}_{2}$ in the case of all weights set to be 1 are $\tau_{1}$ and $\tau_{2}$, in the case of weights set as in~\eqref{weights of inter-plane topology} are $\tilde{\tau}_{1}$ and $\tilde{\tau}_{2}$, respectively.
Then we have:
\begin{align*}
    \tilde{\tau}_{1} &= \tau_{1}\tilde{\xi} + \max_{ \mathcal{P}_{i,j}\in\mathcal{T}_{1}} \sum_{e_{i,j}\in \mathcal{P}_{i,j}}\xi_{i,j} < \tau_{1}\tilde{\xi}+\tilde{\xi} \\
    &\leq  \tau_{2}\tilde{\xi} < \tau_{2}\tilde{\xi} + \max_{ \mathcal{P}_{i,j}\in\mathcal{T}_{2}} \sum_{e_{i,j}\in \mathcal{P}_{i,j}}\xi_{i,j} = \tilde{\tau}_{2},
\end{align*}
where $\mathcal{P}_{i,j}$ is the set of edges representing the shortest path connecting $i$ and $j$. 
Such weight setting essentially guarantees the dominant effect of diameter on searching, and the quality of  communication links determines the optimality in the situation of multiple spanning trees with same diameter.
Therefore, we can reformulate the problem as:
\begin{equation}
    \mathscr{P}_{2}: \quad\quad \underset{\mathcal{T}=\bracket{\mathcal{V}, \tilde{\mathcal{E}}}}{\text{minimize}}  \quad \max_{\mathcal{P}_{i,j} \in \mathcal{G}}\sum_{e_{i,j}\in \mathcal{P}_{i,j}}w\bracket{e_{i,j}}.
\end{equation}

To solve this problem, we employ the idea in~\cite{hassin1995minimum} to treat it as an \emph{absolute 1-center problem} (A1CP) based on the proved equivalence between them.
Specifically, let $A\bracket{\mathcal{G}}$ be the continuum set of points on the edges of $\mathcal{G}$, and $d_{\mathcal{G}}\bracket{i,j}:=\min_{\mathcal{P}_{i,j}\in\mathcal{G}}\sum_{e\in \mathcal{P}_{i,j}}w\bracket{e}$ represents the length of a shortest path in $A\bracket{\mathcal{G}}$ connecting $i$ and $j$.
The goal of A1CP is to minimize the function:
\begin{equation}
    \mathcal{C}\bracket{i} = \max_{j\in\mathcal{V}}d_{\mathcal{G}}\bracket{i,j},
\end{equation}
which is essentially to find the point that minimizes the radius over $\mathcal{G}$.
The core idea of the solution is quite simple, we just need to iterate all points in $A\bracket{\mathcal{G}}$, and for each point $i$, construct a shortest path tree $\mathcal{T}_{i}$ by augmenting $i$ to all vertices in $\mathcal{V}$ based on the shortest path between them.
In the end, the tree corresponds to the optimal point is exactly the inter-plane routing tree that are we looking for.
The overall procedure for inter-plane routing tree optimization is summarized in Algorithm~\ref{topology optimization}.

\begin{algorithm}
    \caption{Proposed Inter-Plane Routing Tree Optimization Algorithm}
    \label{topology optimization}
    \textbf{Initialization}: inter-plane connectivity graph $\mathcal{G} = \bracket{\mathcal{V}, \mathcal{E}}$.\\
    \Foreach{edge $e_{i,j} \in \mathcal{E}$}{
        calculate the communication quality between orbits as in~\eqref{communication quality}.\\
        set the weight for the edge as in~\eqref{weights of inter-plane topology}.
    }
    construct the continuum set of points $A(\mathcal{G})$.\\
    \Foreach{vertex $i \in A(\mathcal{G})$}{
        \Foreach{vertex $j \in \mathcal{V}$ except $i$}{
            calculate the shortest path $\mathcal{P}_{i,j}$.
        }
        construct the shortest path tree $\mathcal{T}_{i}$ for vertex $i$.
    }
    find the vertex $i^{*}$ that minimizes $\mathcal{C}\bracket{i}$. \\
    \textbf{Return}: $\mathcal{T}_{i^{*}}$.
\end{algorithm}

\section{Experimental Results}
\label{simulation results section}
In this section, we shall evaluate the performance of the proposed method with respect to inter-plane aggregation approach, brain-inspired models, and system optimization. 
To begin with, we first present the experiment settings.
As for the learning task, we consider the land cover classification on EuroSAT dataset, which consists of 27000 remote sensing RGB images captured by Sentinel-2 satellites.
The training data is assumed to be non-IID among orbit planes, and we use a Dirichlet process to partition them, together with a parameter $\varsigma$ to control the heterogeneity. 
As for the learning model, we adopt the following two spiking models throughout the simulation:
\begin{itemize}
    \item \textbf{Spiking CNN}, which consists of a convolutional layer with four $3 \times 3$ kernels, a convolutional layer with eight $3 \times 3$ kernels, a fully connected layer and the final classification layer.
    \item \textbf{Spiking Resnet}, for the sake of convenience, we reduce the basic blocks employed. It contains a convolutional layer with sixteen $3 \times 3$ kernels, which followed by three layers consist of 1, 2, 2 basic blocks, respectively.
\end{itemize}
The architectures of the above spiking models maintain the same as the corresponding traditional neural networks, while all the neurons are replaced by the LIF neurons introduced in Section II.
Specifically, the timestamp of spiking models is set as $T=3$, and we adopt $h(\bm{x}) = 1/(1+e^{-\alpha \bm{x}})$ to serve as the surrogate function, where the initial width is set to $\alpha=0.2$, and the mask probability is set to $p=0.2$.

\subsection{Effectiveness of Proposed Inter-Plane Aggregation Scheme}
\begin{figure}[htbp!]
    \centering
    \subfloat[3D view]{
        \label{Walker Star 3D}
        \begin{minipage}[h]{0.2\textwidth}
            \centering
            \includegraphics[width=1\linewidth]{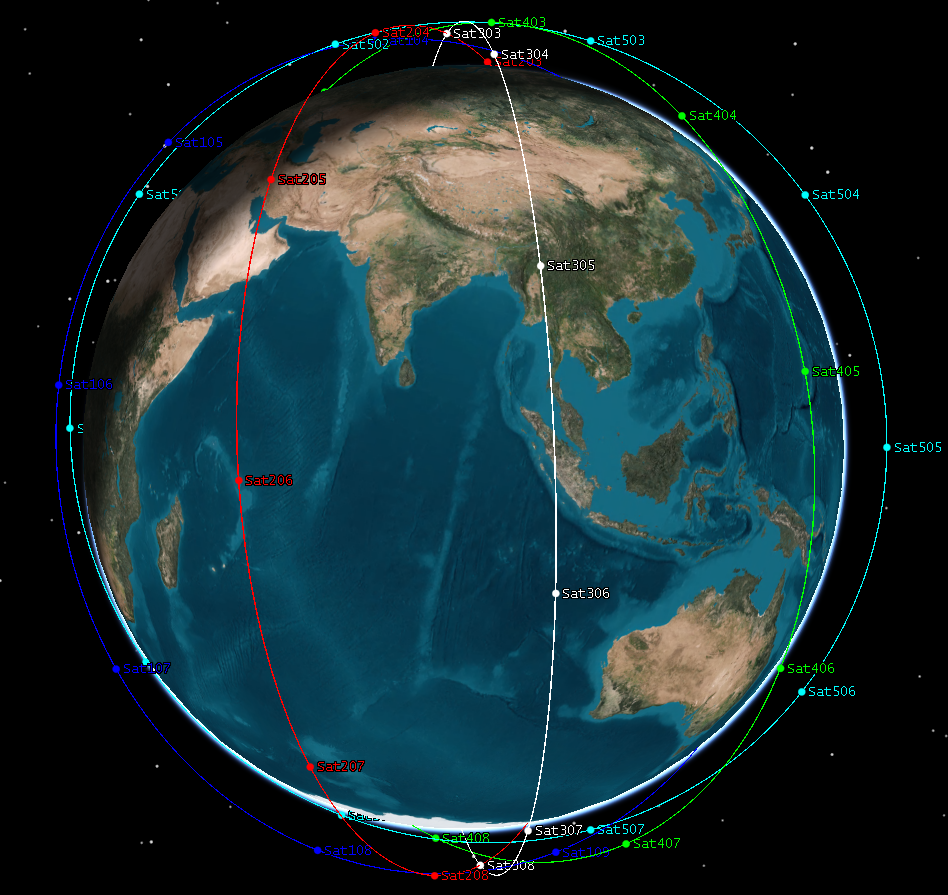}
        \end{minipage}
    }
    \hfill
    \subfloat[2D view]{
        \label{Walker Star 2D}
        \begin{minipage}[h]{0.26\textwidth}
            \centering
            \includegraphics[width=1\linewidth]{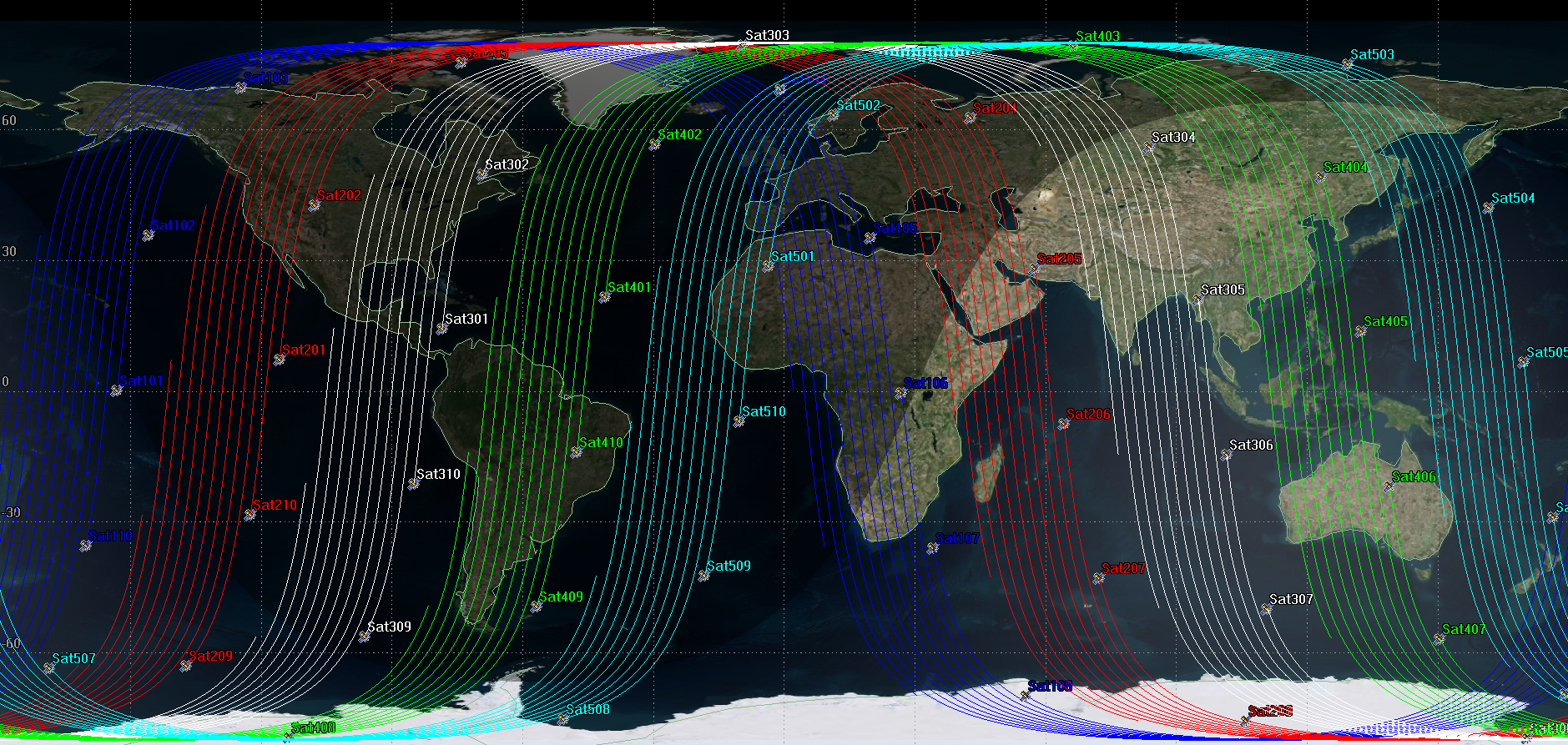}
        \end{minipage}
    }
    \caption{The $50/5/1$ Walker Star Constellation.}
    \label{Walker Star Constellation}
\end{figure}

In this experiment, we evaluate the performance of the proposed RelaySum based inter-plane aggregation approach.
The learning rate is set to $0.05$ and $0.1$ for Spiking CNN and Spiking Resnet, respectively, and the heterogeneity parameter is set as $\varsigma=0.02$.
As for the network settings, we consider the $50/5/1$ Walker Star constellation as illustrated in Fig~\ref{Walker Star Constellation}.
Under such constellation, the planes naturally formulates a chain topology, and we compare the following three different schemes for inter-plane model aggregation:
\begin{itemize}
    \item \textbf{Proposed Algorithm}, where the model updates from different planes are aggregated via RelaySum.
    \item \textbf{Gossip Averaging} \cite{zhou2024decentralized, yang2024dfedsat, yan2023convergence, meng2021decentralized}, where each plane only communicates with its neighbor on the formulated topology, and updates the model via averaged received messages.
    \item \textbf{All Reduce} \cite{zhai2023fedleo, wu2022dsfl}, where each plane diffuses its model updates across the formulated topology, and updates the model by the exact global model. 
    Compared with the former two schemes, it requires several inter-plane communication rounds to finish the model aggregation process.
\end{itemize}
\begin{figure}[htbp!]
    \centering
    \subfloat[Training Loss of Spiking CNN]{
        \label{training loss for cnn}
        \begin{minipage}[t]{0.23\textwidth}
            \centering
            \includegraphics[width=1\linewidth]{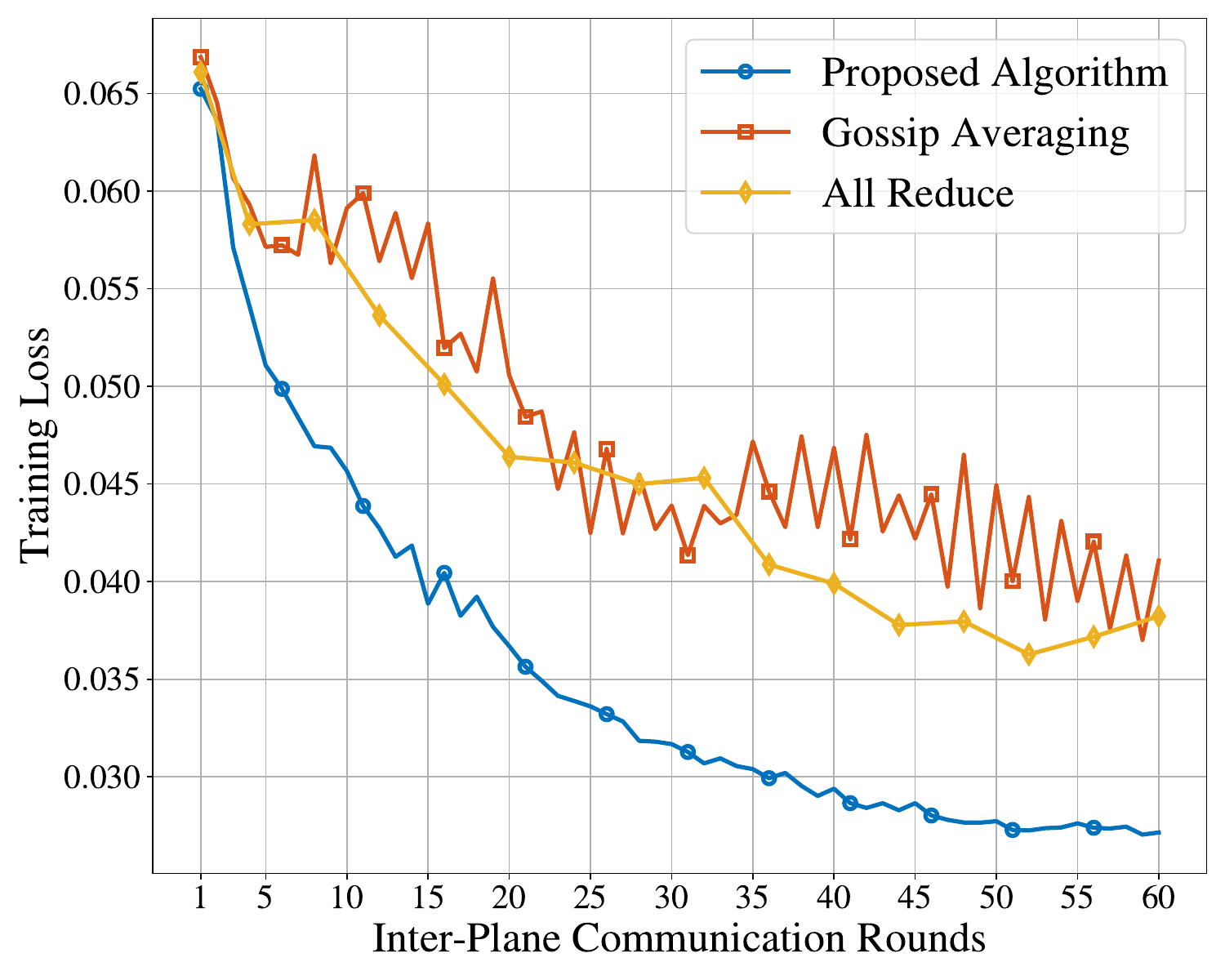}
        \end{minipage}
    }
    \hfill
    \subfloat[Test Accuracy of Spiking CNN]{
        \label{test accuracy for cnn}
        \begin{minipage}[t]{0.23\textwidth}
            \centering
            \includegraphics[width=1\linewidth]{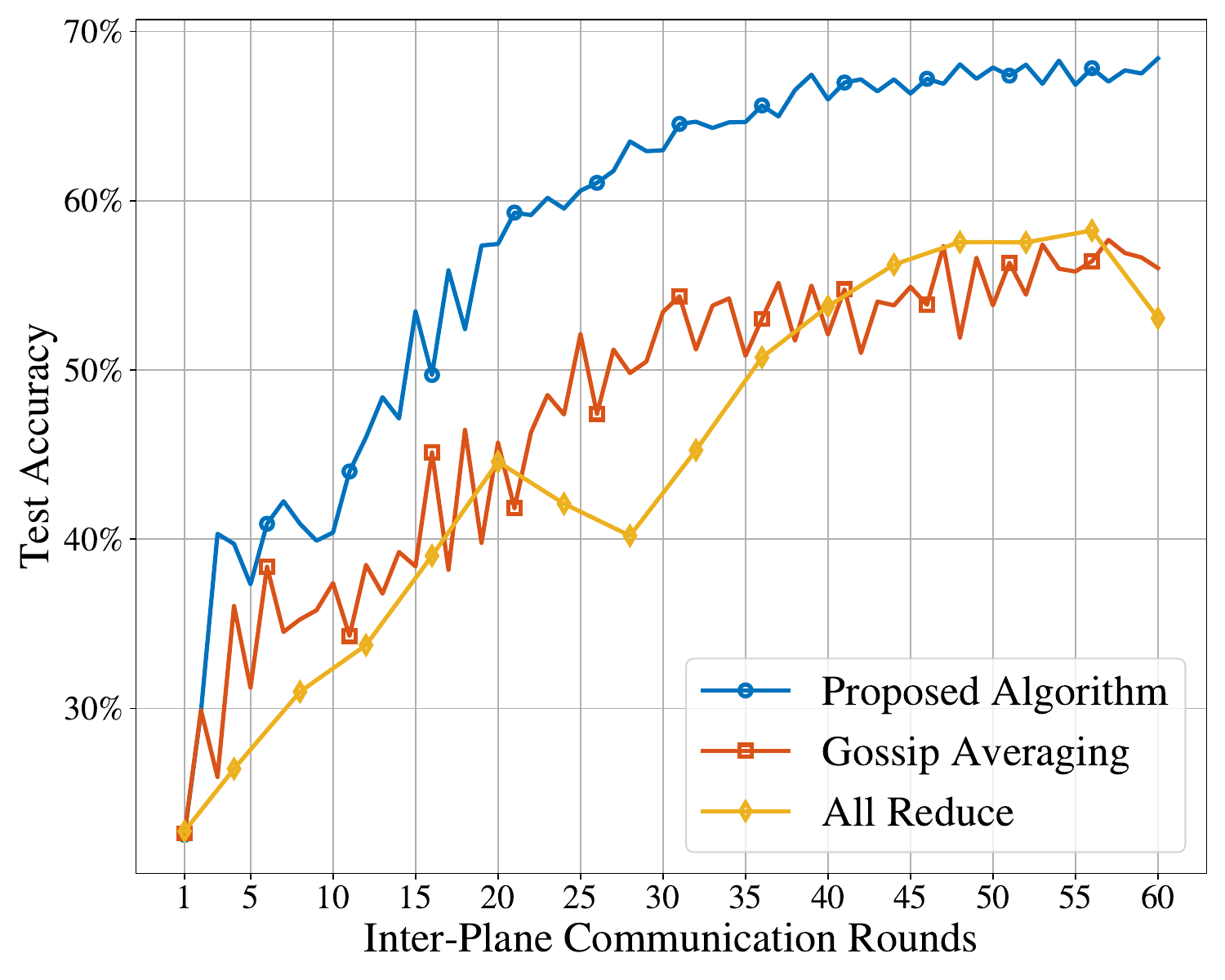}
        \end{minipage}
    }
    \hfill
    \subfloat[Training Loss of Spiking Resnet]{
        \label{training loss for resnet}
        \begin{minipage}[t]{0.23\textwidth}
            \centering
            \includegraphics[width=1\linewidth]{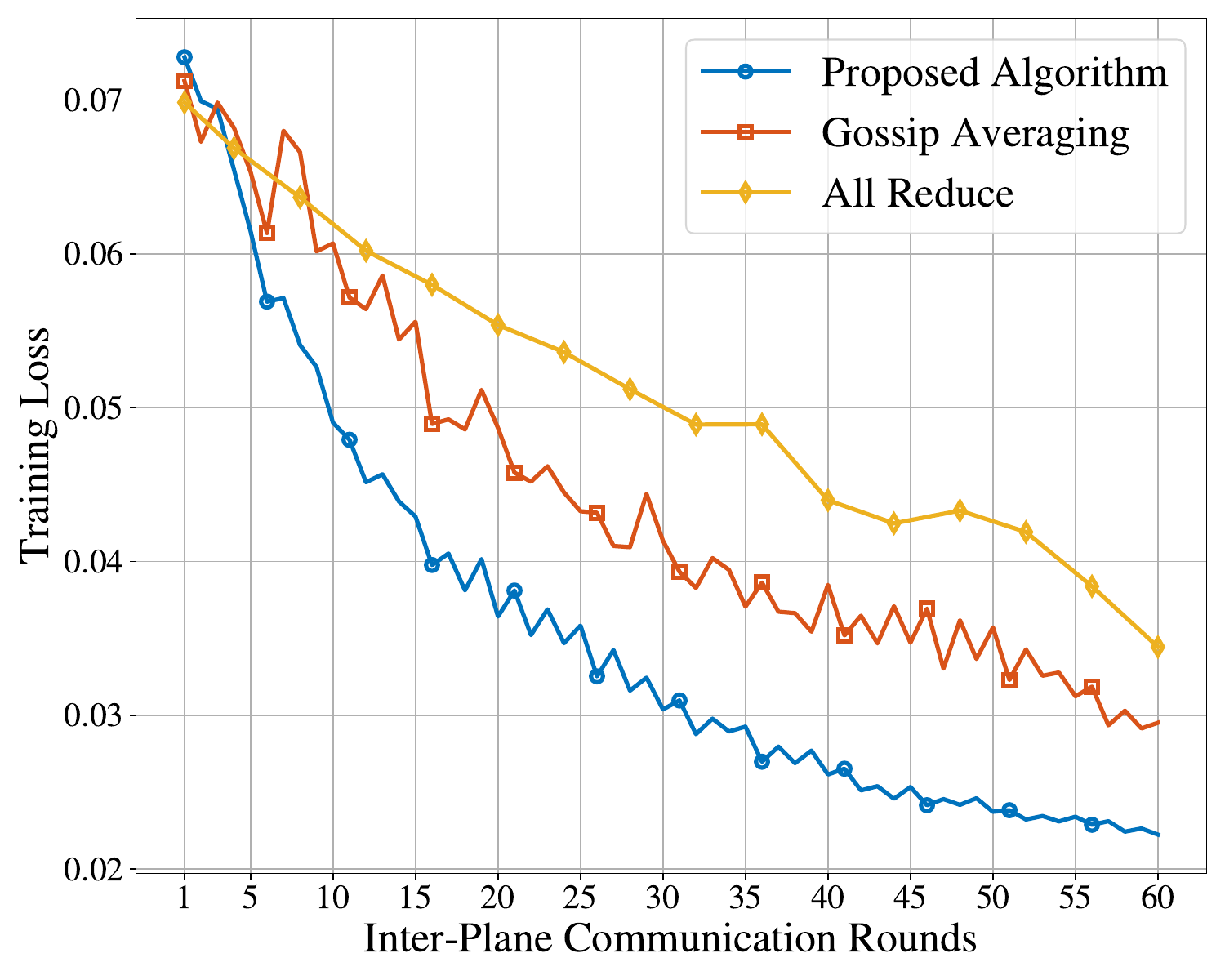}
        \end{minipage}
    }
    \hfill
    \subfloat[Test Accuracy of Spiking Resnet]{
        \label{test accuracy for resnet}
        \begin{minipage}[t]{0.23\textwidth}
            \centering
            \includegraphics[width=1\linewidth]{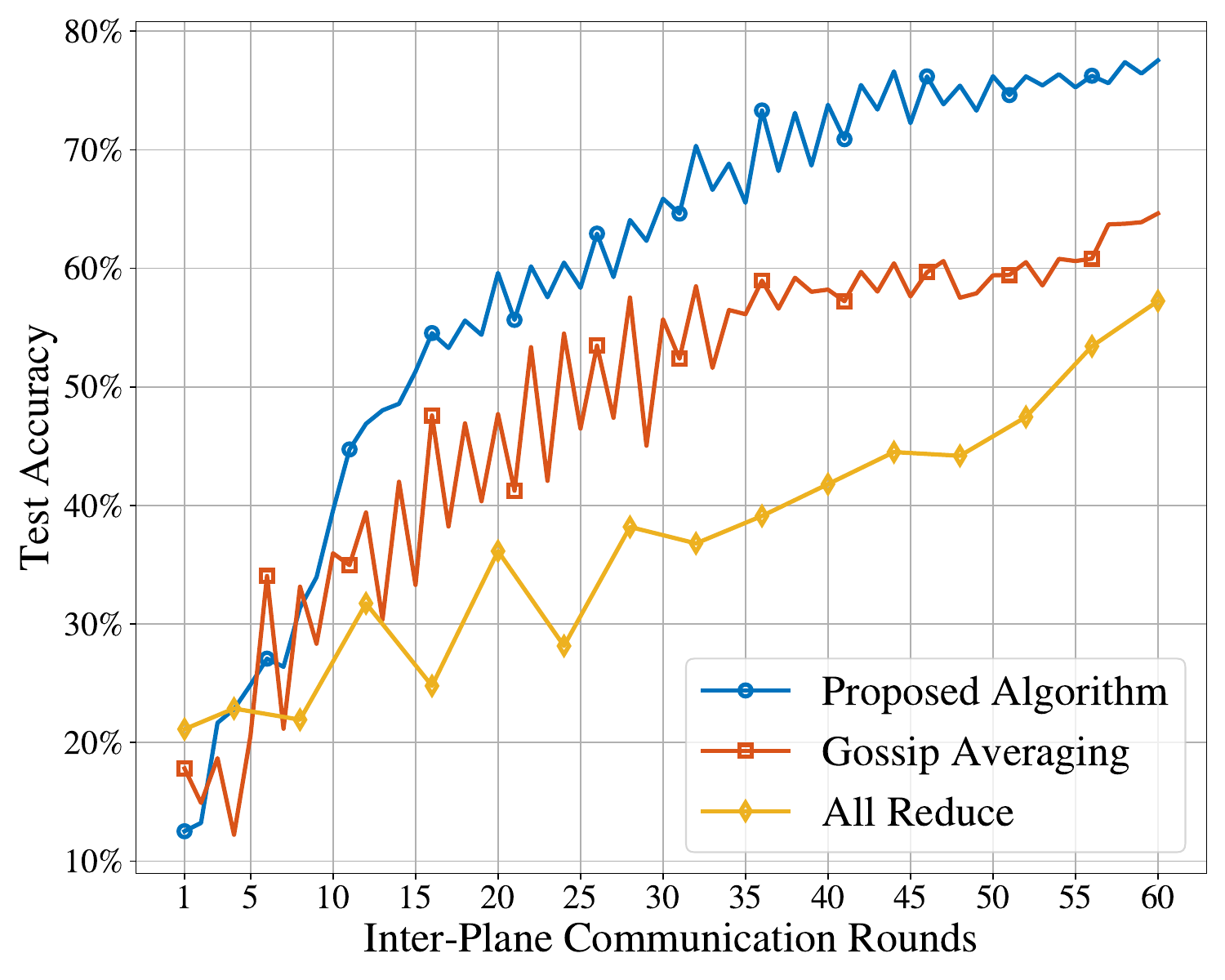}
        \end{minipage}
    }
    \caption{Comparison of different inter-plane model aggregation schemes.}
    \label{inter-plane comparison}
\end{figure}

Fig.~\ref{inter-plane comparison} plots the performance of decentralized satellite neuromorphic learning under different inter-plane model aggregation schemes in terms of training loss and test accuracy, respectively. 
It is observed that for both Spiking CNN and Spiking Resnet, our proposed algorithm is able to reach a small optimality gap and maintain a relatively high test accuracy after $60$ inter-plane communication rounds, while the other two schemes still keep a relatively poor performance (accuracy gap larger than 10\%).
This is due to the drawbacks of these two baseline schemes as we have analyzed in Section II-B.
On the one hand, the repeated weighted averaging in Gossip Averaging makes model updates diffuse slowly on the inter-plane topology, which further leads to a poor convergence rate.
On the other hand, although All Reduce can obtain the exact global model in each iteration, the inter-plane communication rounds required scale with the size of network, which results in the inefficiency of communication.
Comparatively, our proposed RelaySum based inter-plane aggregation only consumes one inter-plane communication rounds per iteration, and the model updates information is fully utilized.
Therefore, it is able to achieve a better learning performance.

\subsection{Energy Efficiency of Employed Brain-Inspired Model}
In this experiment, we compare the performance of the employed neuromorphic models and artificial neural networks.
The learning rate is set to $0.05$ and $0.25$ for CNN and Resnet, respectively, and the heterogeneity parameter is set as $\varsigma=0.2$.
As for the network settings, we adopt the $50/5/1$ Walker Star constellation in the previous experiment.

\begin{figure}[htbp!]
    \centering
    \subfloat[Training Loss of CNN]{
        \begin{minipage}[t]{0.23\textwidth}
            \centering
            \includegraphics[width=1\linewidth]{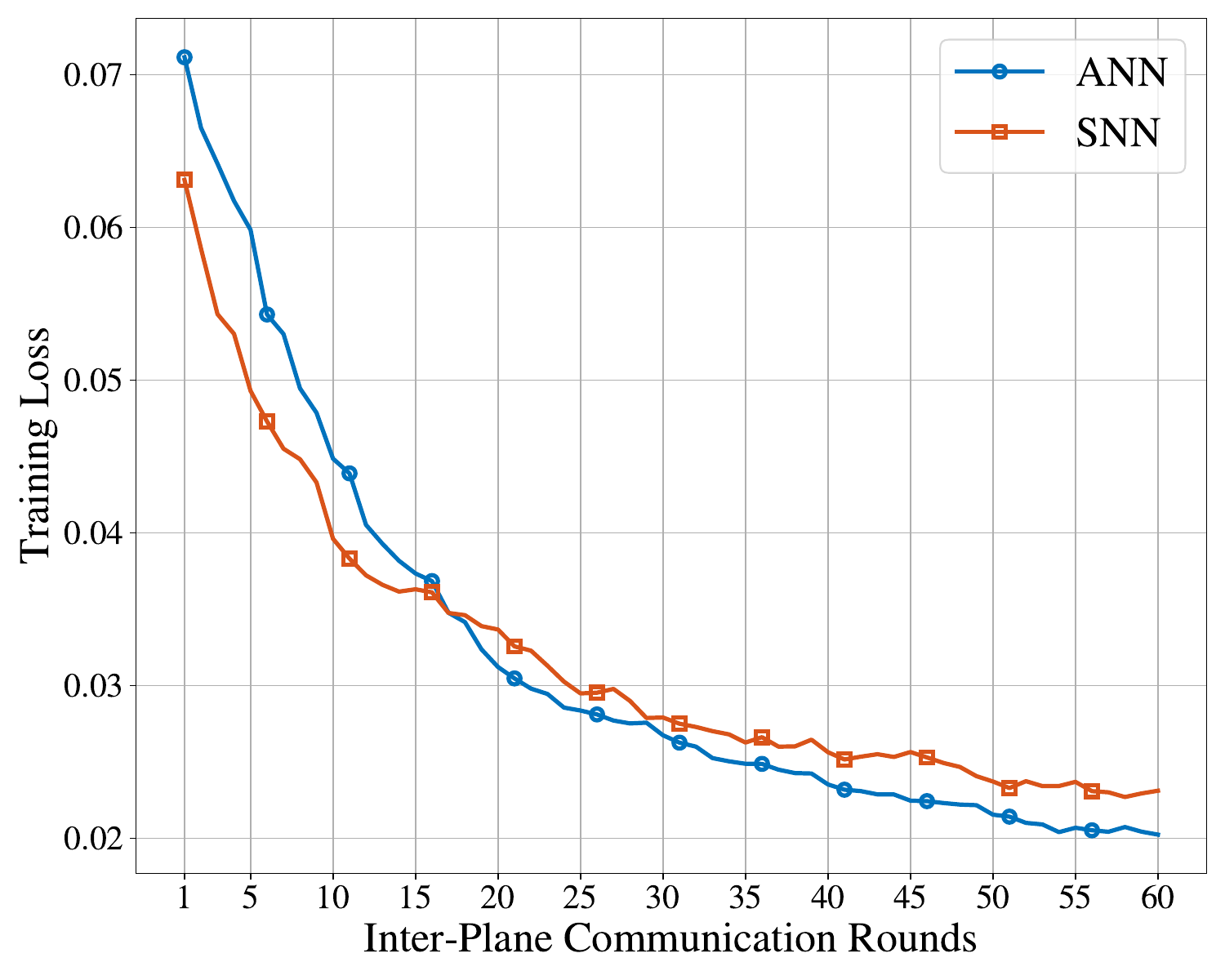}
        \end{minipage}
    }
    \hfill
    \subfloat[Test Accuracy of CNN]{
        \begin{minipage}[t]{0.23\textwidth}
            \centering
            \includegraphics[width=1\linewidth]{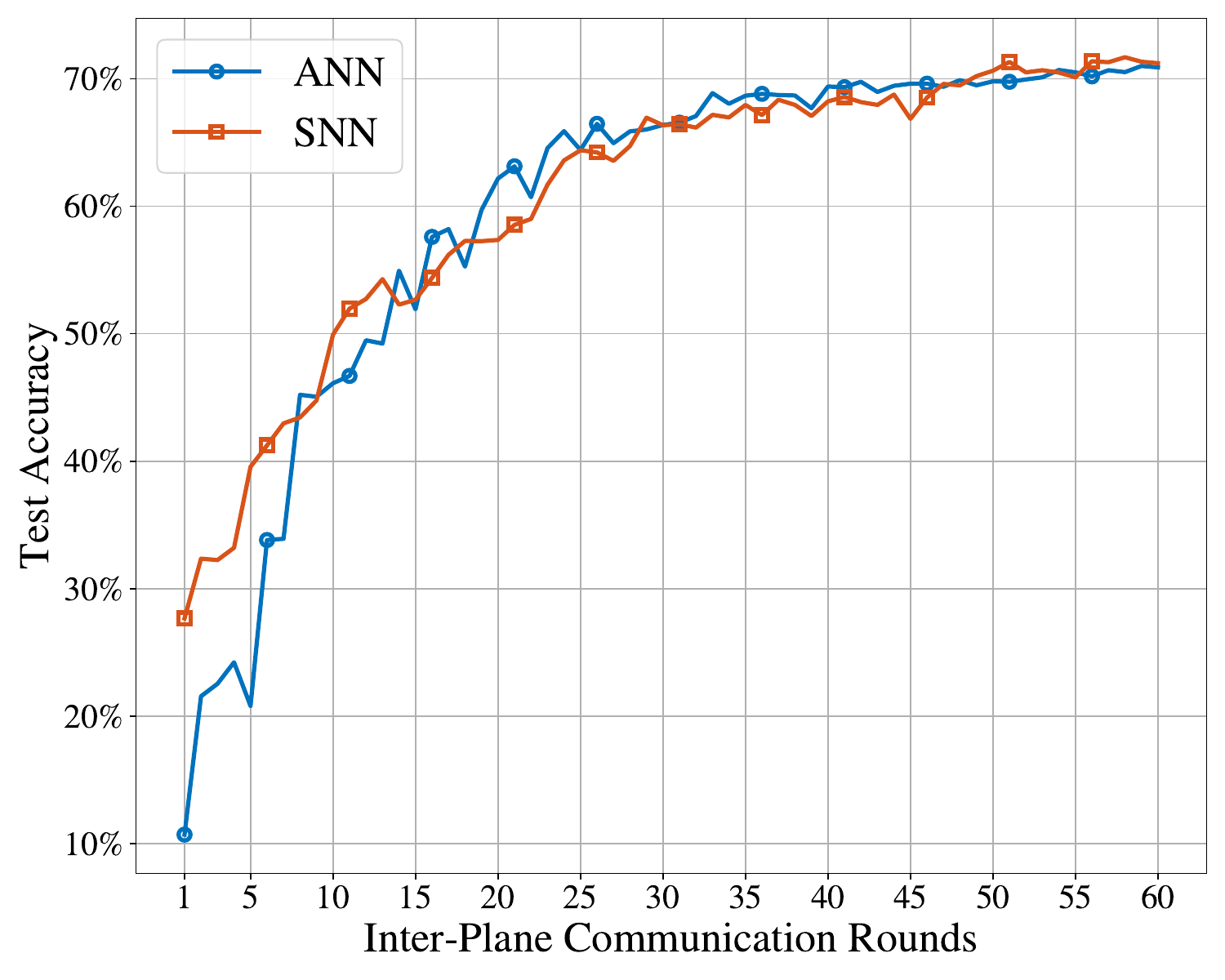}
        \end{minipage}
    }
    \hfill
    \subfloat[Training Loss of Resnet]{
        \begin{minipage}[t]{0.23\textwidth}
            \centering
            \includegraphics[width=1\linewidth]{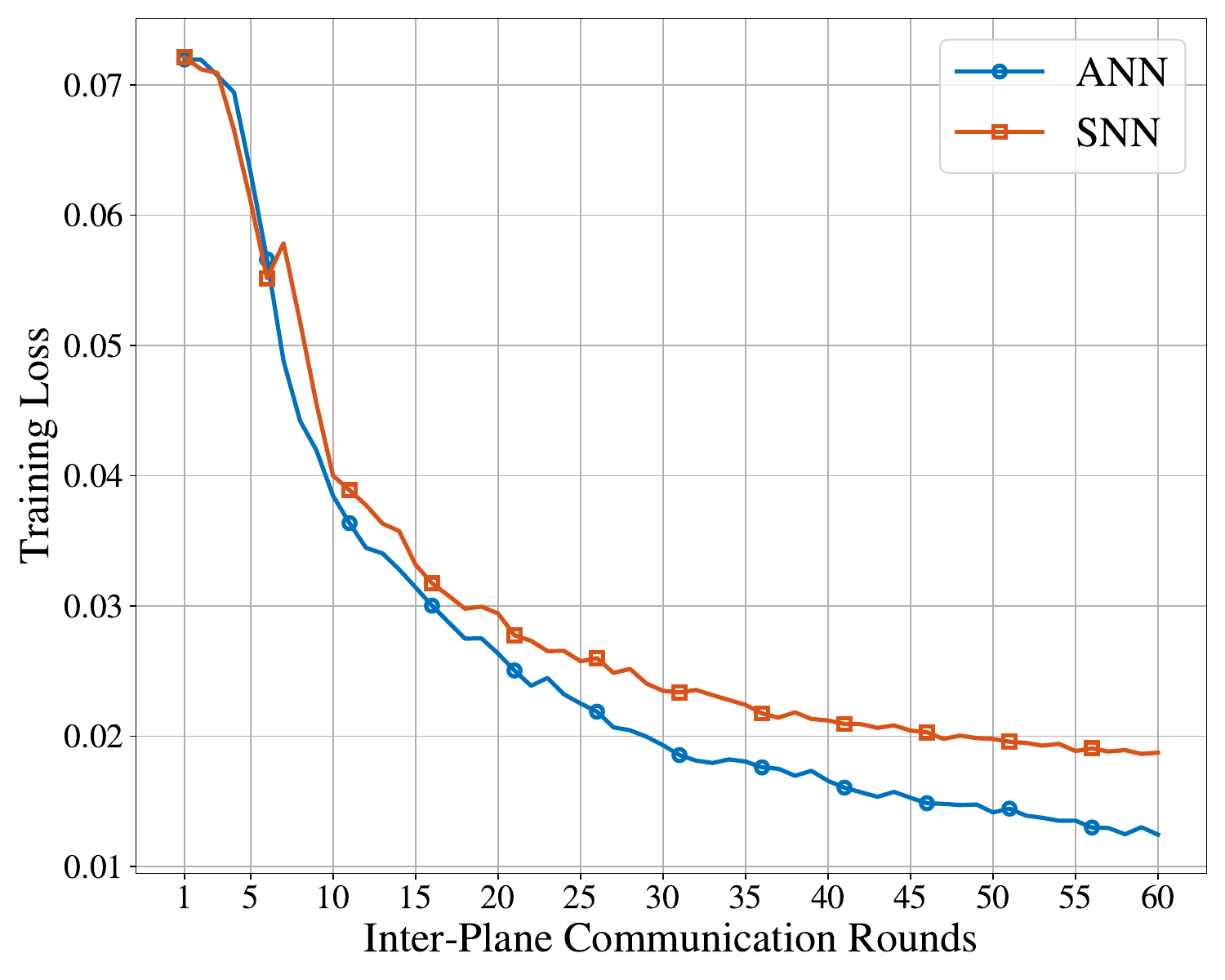}
        \end{minipage}
    }
    \hfill
    \subfloat[Test Accuracy of Resnet]{
        \begin{minipage}[t]{0.23\textwidth}
            \centering
            \includegraphics[width=1\linewidth]{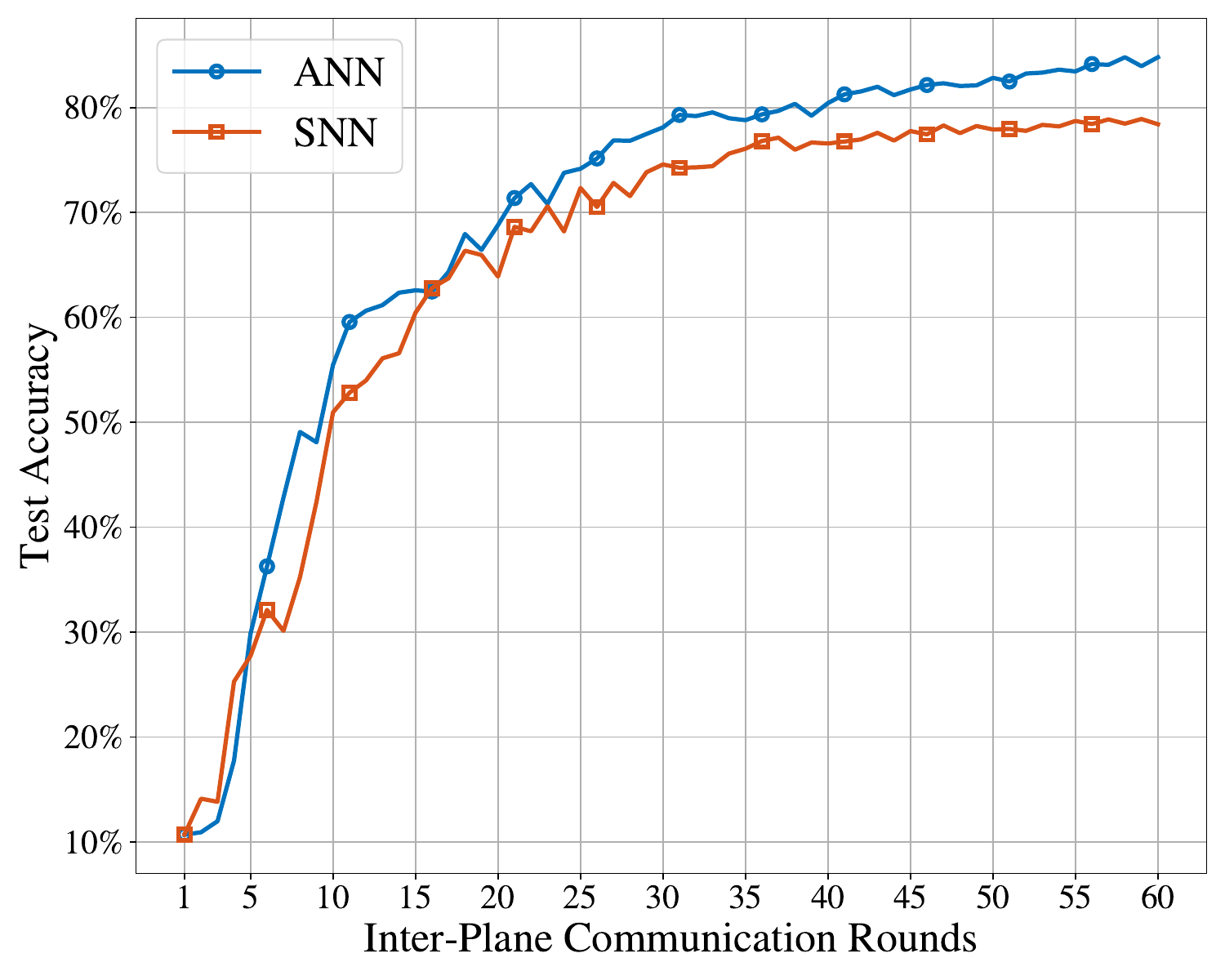}
        \end{minipage}
    }
    \caption{Learning performance comparison of ANN and SNN.}
    \label{ANN-SNN comparison}
\end{figure}

We first consider the learning performance, Fig.~\ref{ANN-SNN comparison} shows the decentralized satellite learning performance comparison of ANN and SNN in terms of training loss and test accuracy, respectively.
It can be seen that both ANN and SNN reach a low loss value and maintain a high precision level quickly, and the performance difference between ANN and SNN is very small (less than 3\% accuracy gap). 
Therefore, the employed brain-inspired models have a good learning performance.

\begin{figure}[htbp!]
    \centering
    \subfloat[Spiking Rates of Spiking CNN]{
        \begin{minipage}[t]{0.23\textwidth}
            \centering
            \includegraphics[width=1\linewidth]{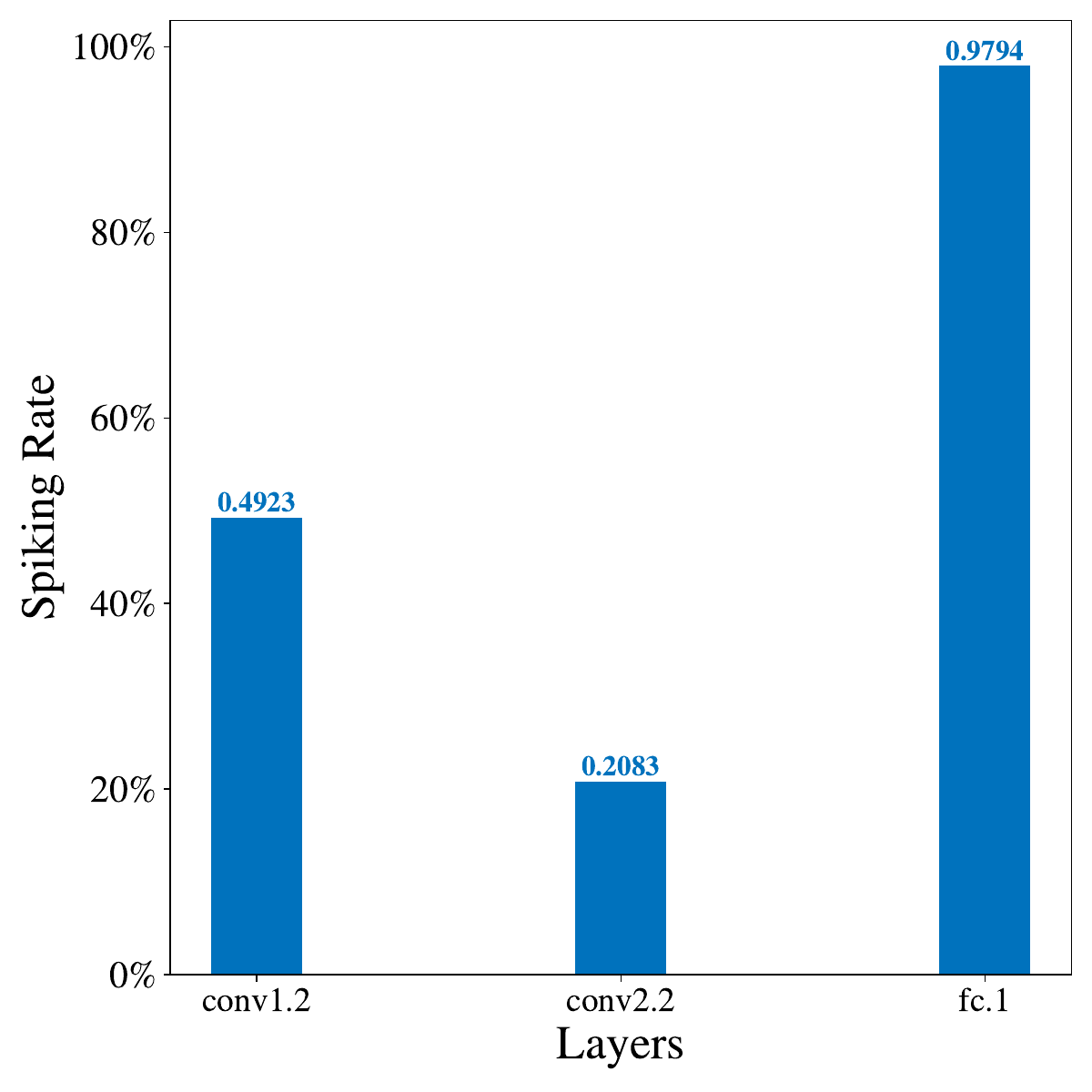}
        \end{minipage}
    }
    \hfill
    \subfloat[Energy Consumption of CNN]{
        \begin{minipage}[t]{0.23\textwidth}
            \centering
            \includegraphics[width=1\linewidth]{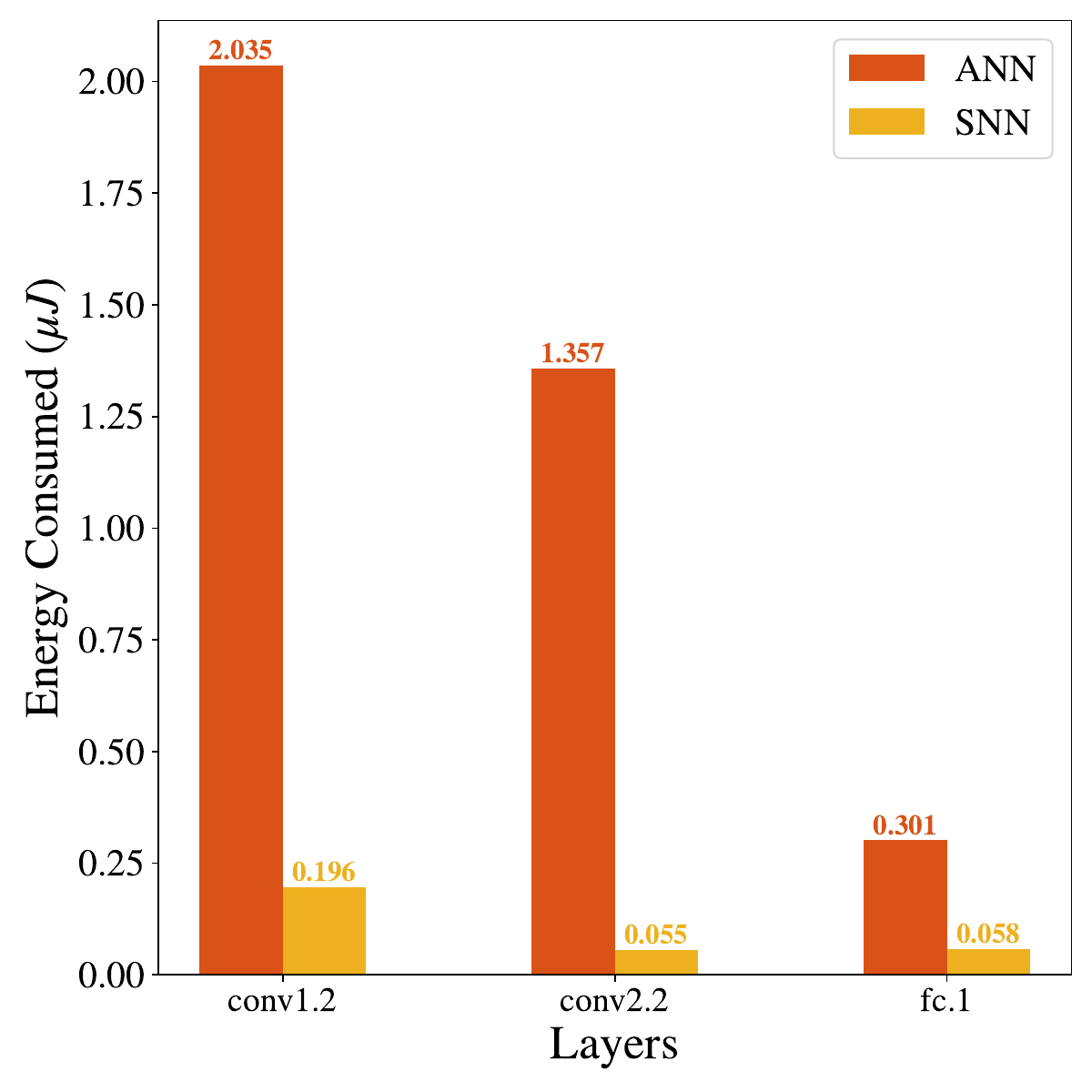}
        \end{minipage}
    }
    \caption{CNN energy consumption comparison of ANN and SNN.}
    \label{Energy comparison CNN}
\end{figure}

\begin{figure}[htbp!]
    \centering
    \subfloat[Spiking Rates of Spiking Resnet]{
        \begin{minipage}[t]{0.46\textwidth}
            \centering
            \includegraphics[width=1\linewidth]{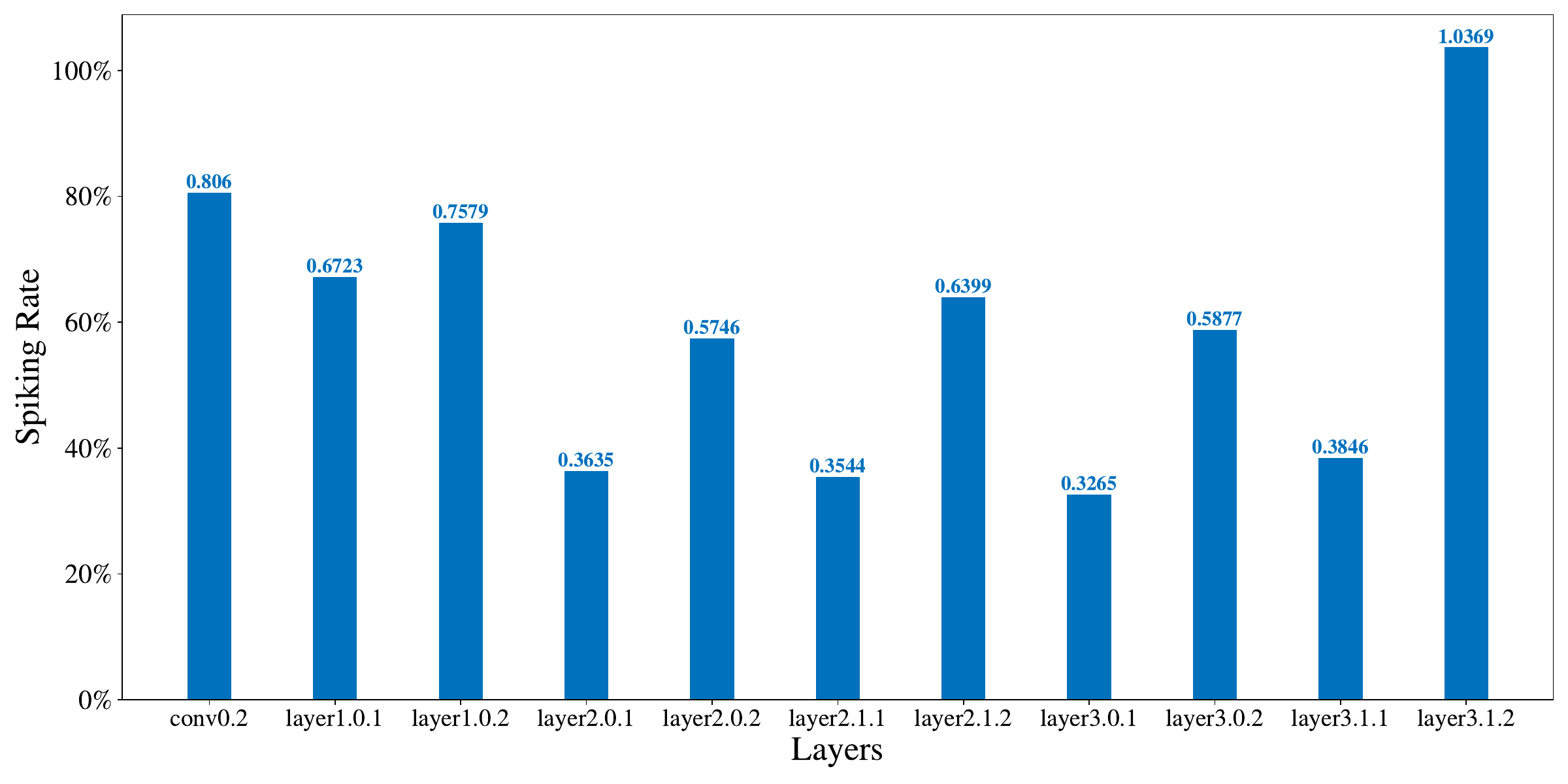}
        \end{minipage}
    }
    \hfill
    \subfloat[Energy Consumption of Resnet]{
        \begin{minipage}[t]{0.46\textwidth}
            \centering
            \includegraphics[width=1\linewidth]{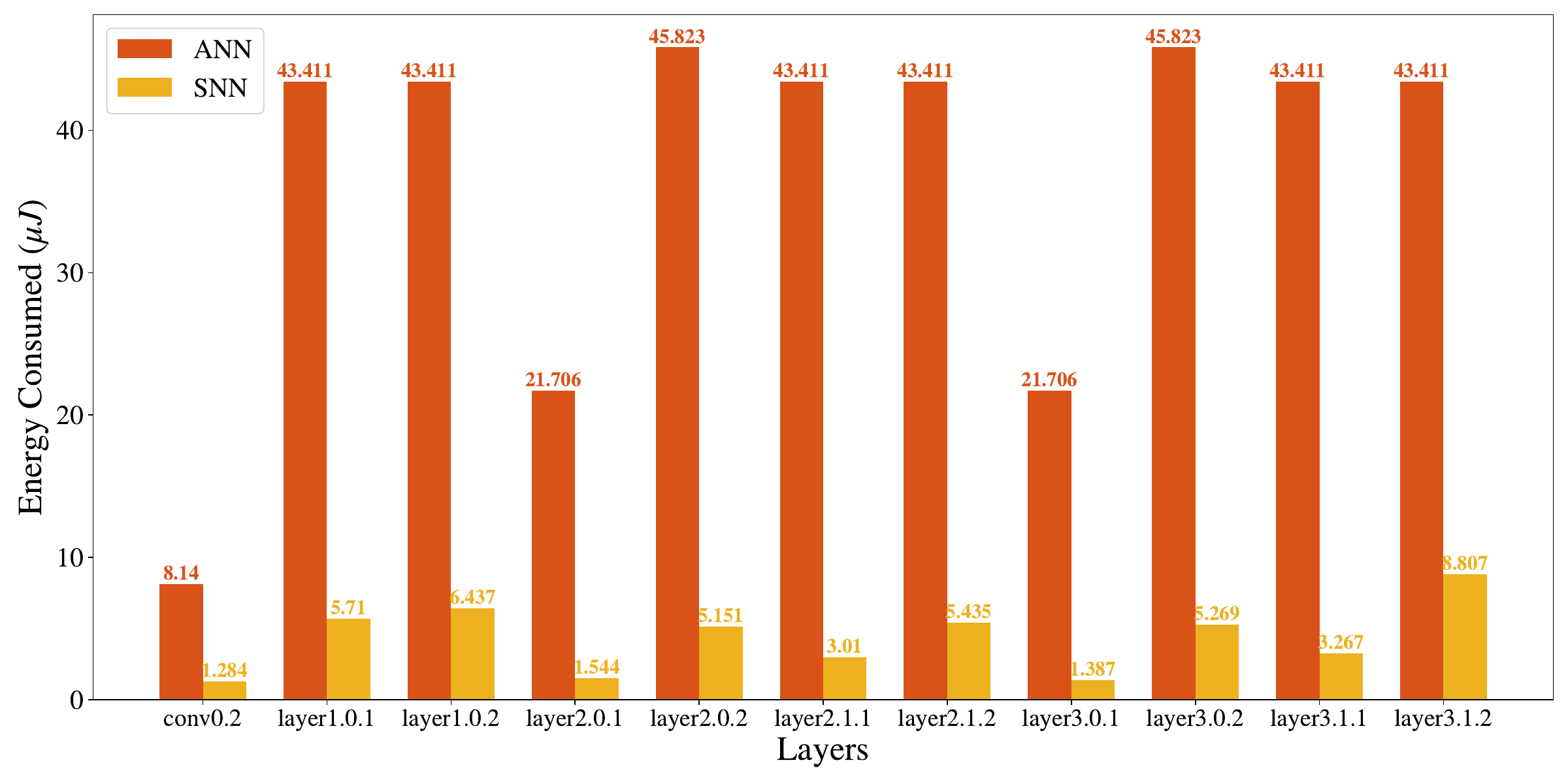}
        \end{minipage}
    }
    \caption{Resnet energy consumption comparison of ANN and SNN.}
    \label{Energy comparison Resnet}
\end{figure}

Then we turn into the energy consumption comparison between ANN and SNN.
Specifically, the energy cost is measured in $45$nm CMOS technology, where each MAC operation in ANN consumes $4.6pJ$ energy and each accumulation operation in SNN consumes $0.9pJ$ energy~\cite{horowitz20141, li2021differentiable}.
The specific number of operations of each layer can be computed according to the corresponding model architecture.
Particularly, due to the sparsity property, the number of operations for SNN also depend on the spiking rates~\cite{rathi2021diet}.
Fig.~\ref{Energy comparison CNN} and Fig.~\ref{Energy comparison Resnet} plot the spiking rates of neuromorphic models and energy consumption comparison between ANN and SNN.
It is observed that for both CNN and Resnet, the energy consumed in each layer of SNN is much more less than that of ANN (10$\times$ reduction in average). 
This is due to the extreme sparsity in computation of SNN, which is able to avoid heavy floating point matrix multiplications in neural networks.
Overall, the employed neurmorphic models enjoy a high energy efficiency and maintain a good learning performance at the same time, which naturally suits the low-power environment in space networks. 

\subsection{Effectiveness of System Optimization}
\begin{figure}[htbp!]
    \centering
    \subfloat[3D view]{
        \label{Walker Delta 3D}
        \begin{minipage}[h]{0.2\textwidth}
            \centering
            \includegraphics[width=1\linewidth]{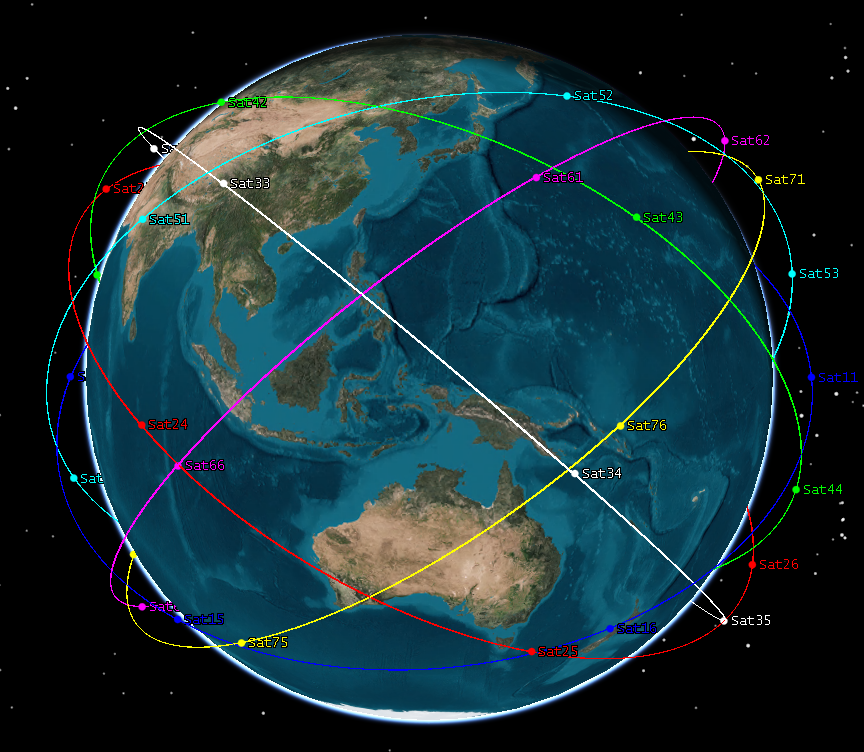}
        \end{minipage}
    }
    \hfill
    \subfloat[2D view]{
        \label{Walker Delta 2D}
        \begin{minipage}[h]{0.26\textwidth}
            \centering
            \includegraphics[width=1\linewidth]{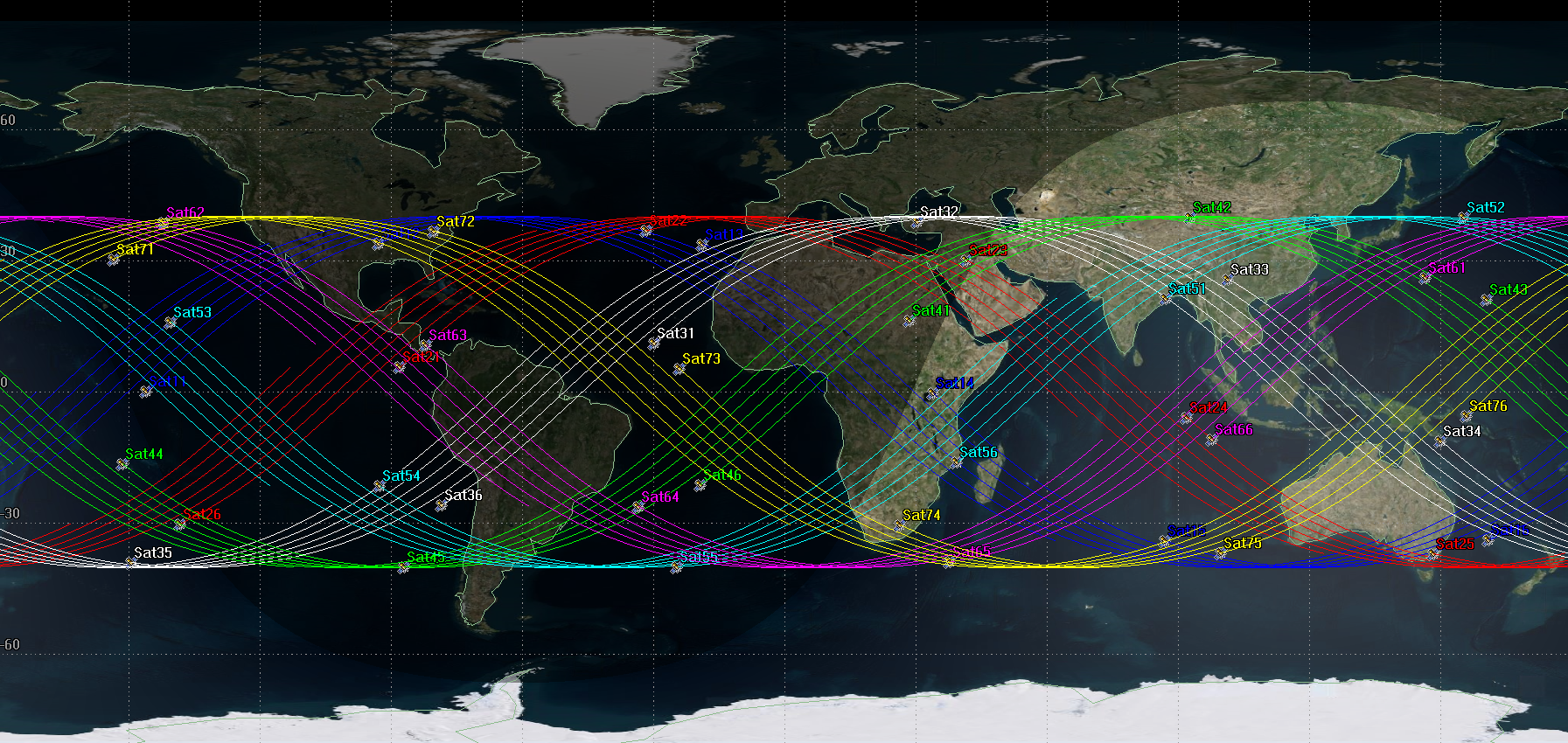}
        \end{minipage}
    }
    \caption{The $42/7/1$ Walker Delta Constellation.}
    \label{Walker Delta Constellation}
\end{figure}

\begin{figure}[htbp!]
    \centering
    \includegraphics[width=0.95\linewidth]{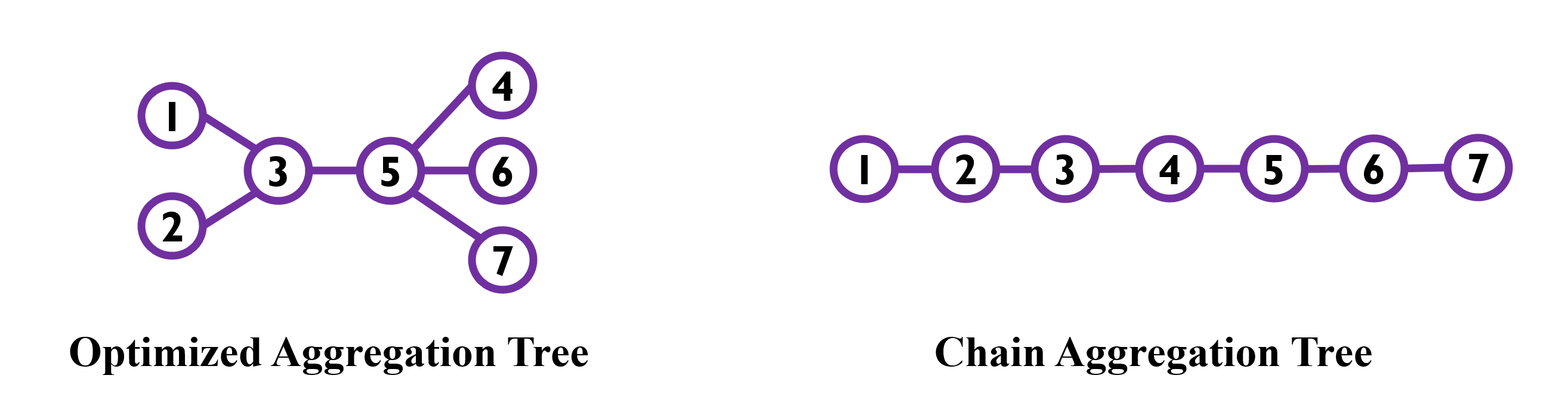}
    \caption{Inter-plane aggregation trees.}
    \label{inter-plane aggregation trees}
\end{figure}

In this experiment, we evaluate the performance of the proposed optimization approach on the inter-plane routing tree for model aggregation.
The learning rate is set to $0.05$ and $0.1$ for Spiking CNN and Spiking Resnet, respectively, and the heterogeneity parameter is set as $\varsigma=1$.
As for satellite communication parameters, the carrier frequency $f_{c}$ is set as $2.4$ GHz, EIRPG is set as $10$ dbW, system bandwidth is set as $32$ MHz, maximum Doppler shift is set as $60$ kHz. 
Specifically, we compare the following two schemes for inter-plane aggregation tree:
\begin{itemize}
    \item \textbf{Optimized Aggregation Tree}, where the inter-plane routing tree for model aggregation is optimized via the proposed Algorithm~\ref{topology optimization}.
    \item \textbf{Chain Aggregation Tree}, where the inter-plane routing tree for model aggregation is formulated by simply searching a chain that connects all planes.
\end{itemize}
As for network settings, we adopt the $42/7/1$ Walker Delta constellation as shown in Fig.~\ref{Walker Delta Constellation}.
Based on this constellation, we can obtain the corresponding aggregation trees in Fig.~\ref{inter-plane aggregation trees}.

\begin{figure}[htbp!]
    \centering
    \subfloat[Training Loss of Spiking CNN]{
        \begin{minipage}[t]{0.23\textwidth}
            \centering
            \includegraphics[width=1\linewidth]{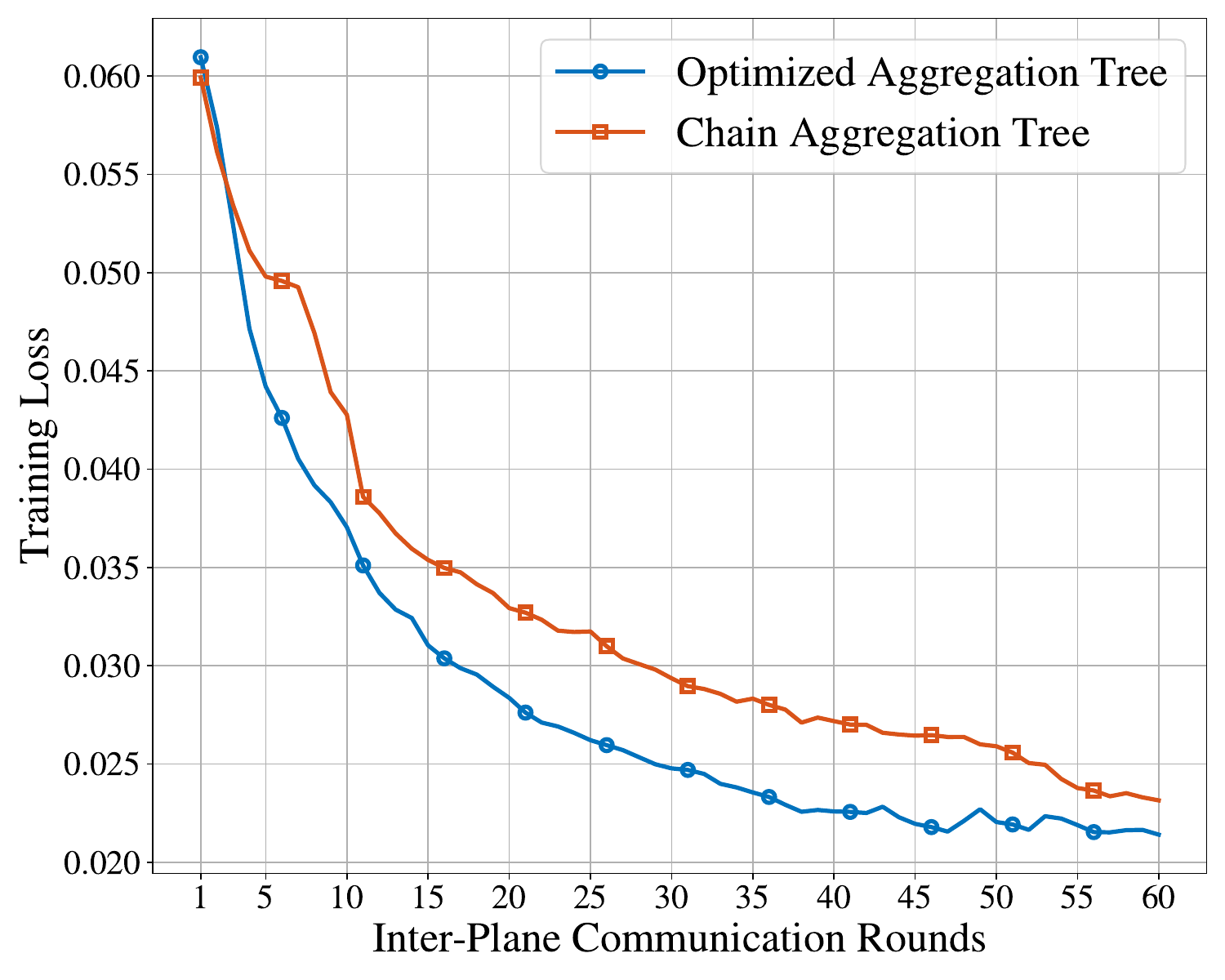}
        \end{minipage}
    }
    \hfill
    \subfloat[Test Accuracy of Spiking CNN]{
        \begin{minipage}[t]{0.23\textwidth}
            \centering
            \includegraphics[width=1\linewidth]{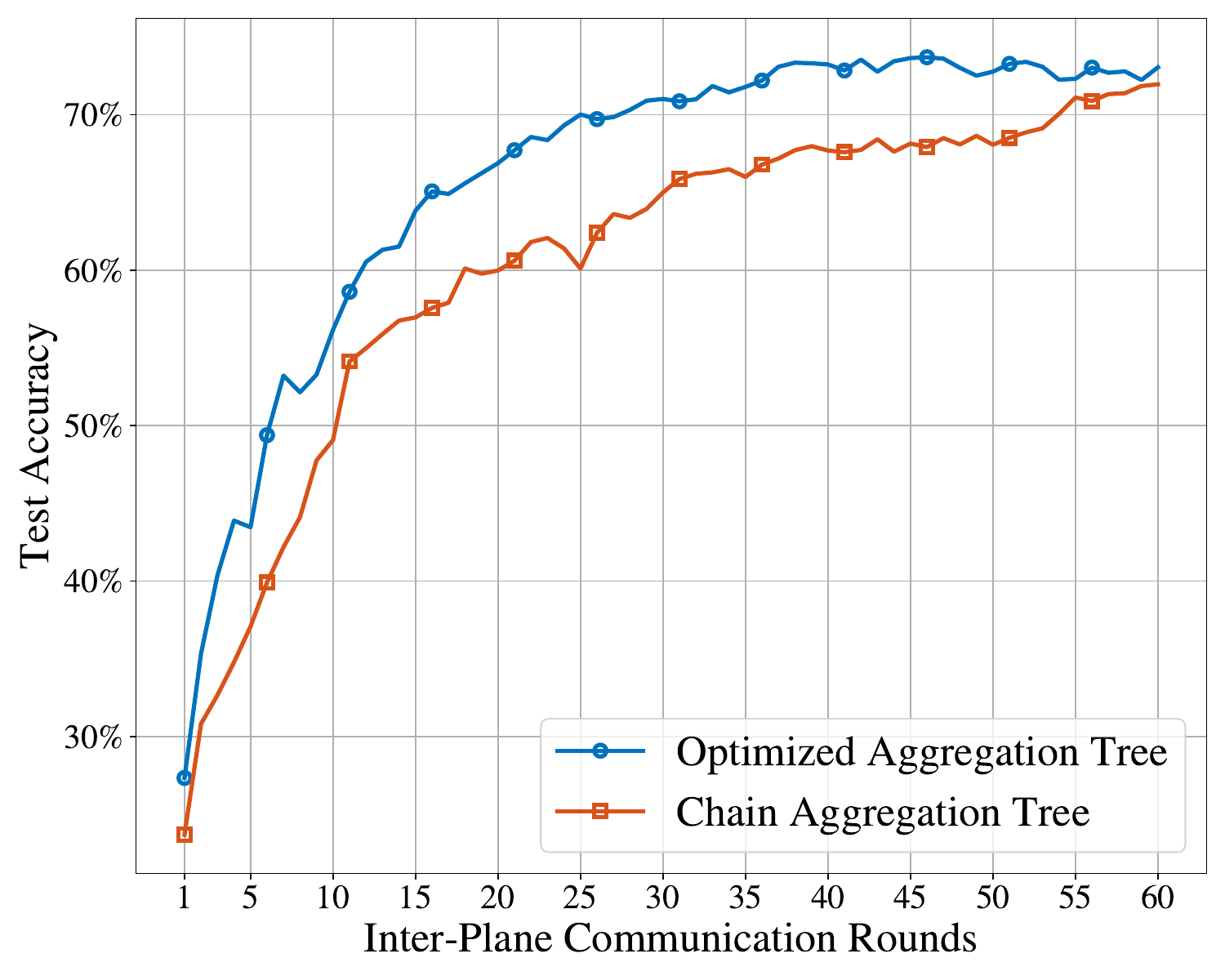}
        \end{minipage}
    }
    \hfill
    \subfloat[Training Loss of Spiking Resnet]{
        \begin{minipage}[t]{0.23\textwidth}
            \centering
            \includegraphics[width=1\linewidth]{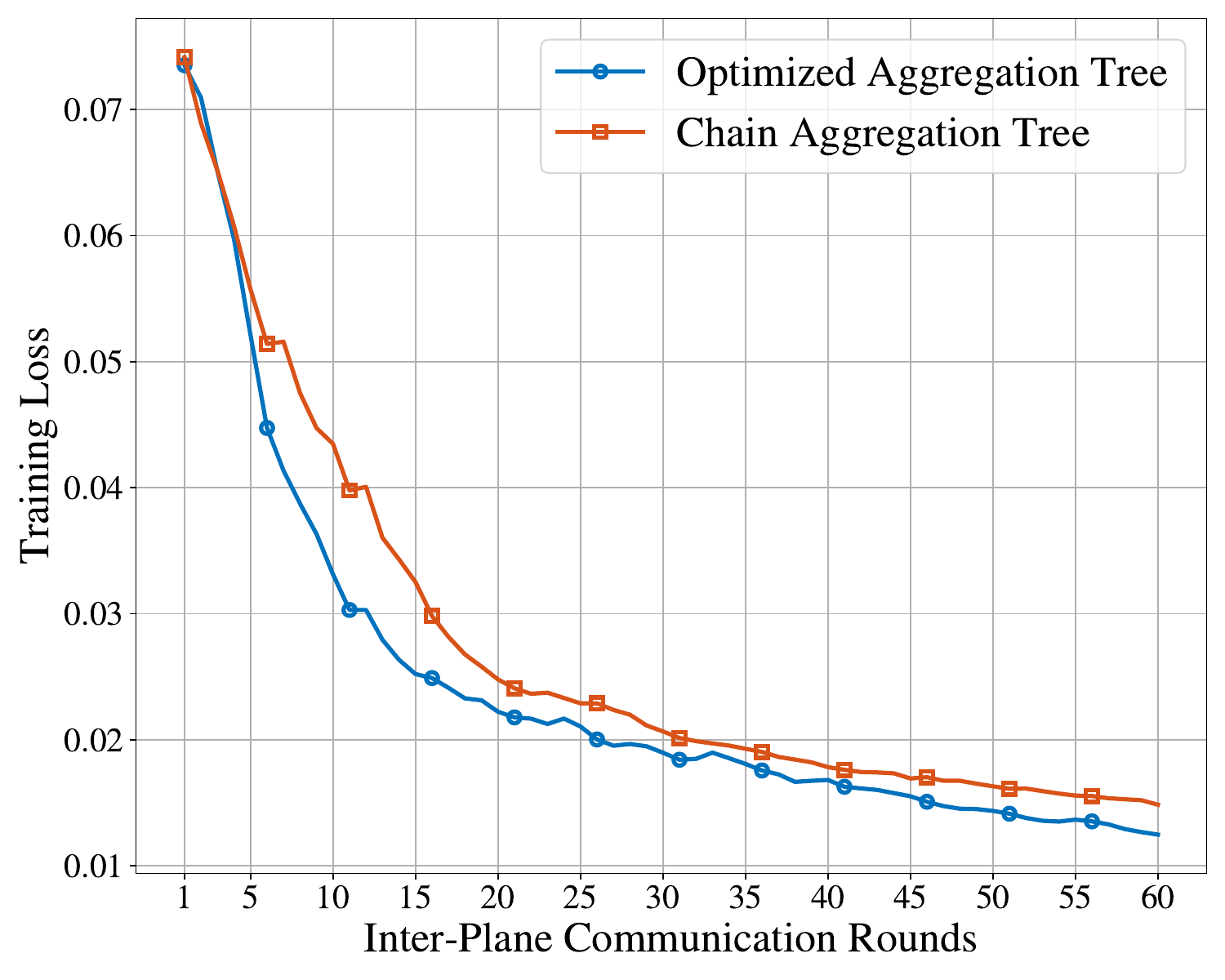}
        \end{minipage}
    }
    \hfill
    \subfloat[Test Accuracy of Spiking Resnet]{
        \begin{minipage}[t]{0.23\textwidth}
            \centering
            \includegraphics[width=1\linewidth]{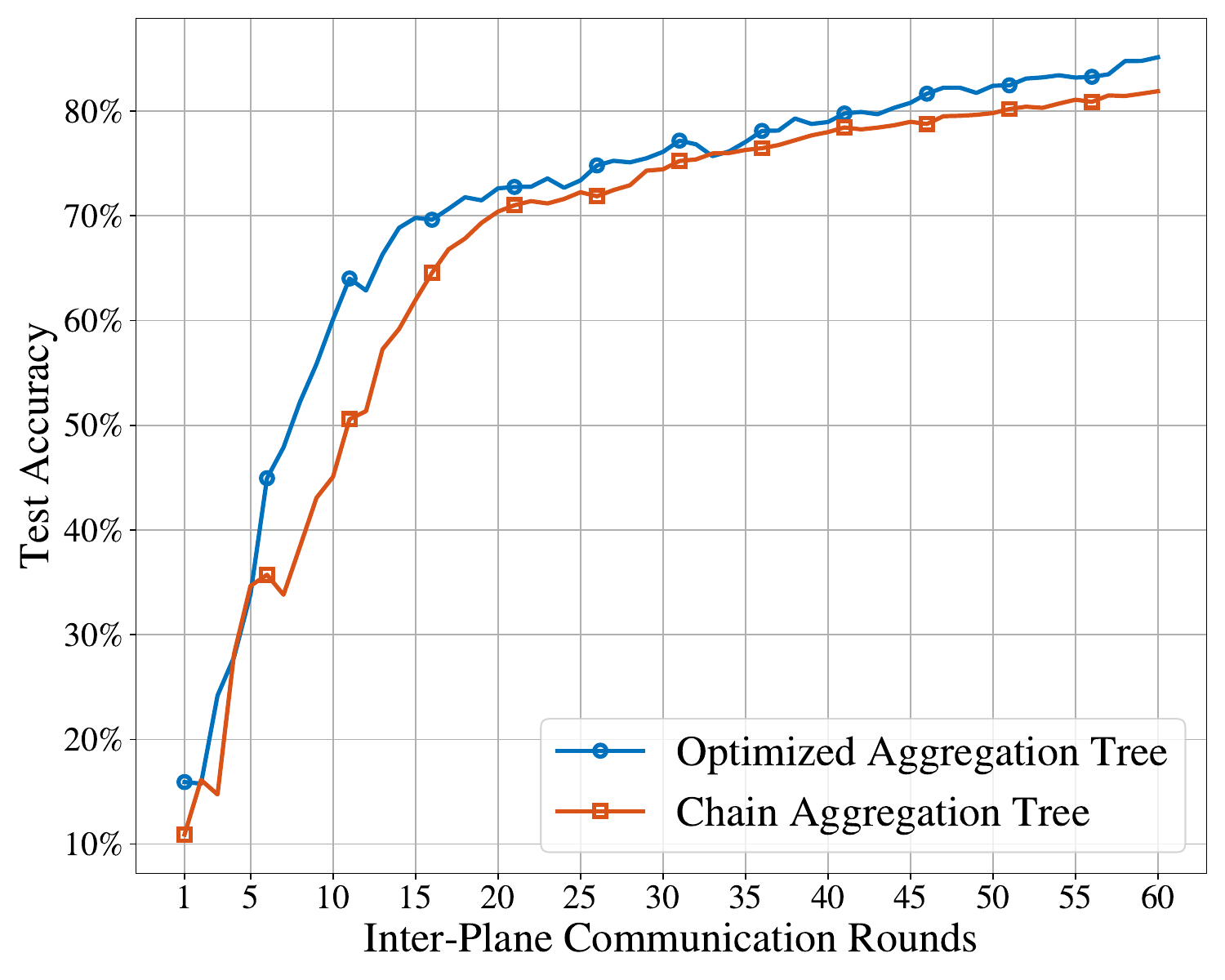}
        \end{minipage}
    }
    \caption{Comparison of different inter-plane aggregation trees.}
    \label{aggregation tree comparison}
\end{figure}

Fig.~\ref{aggregation tree comparison} shows the learning performance under different inter-plane aggregation trees with respect to training loss and test accuracy.
It is observed that for both optimization aggregation tree and chain aggregation tree, the training loss reaches a small optimality gap and the test accuracy keep a high precision level after $60$ inter-plane communication rounds.
Nevertheless, the optimized aggregation tree exhibits a faster convergence rate.
This is because the optimized aggregation tree has a smaller diameter $3$ compared with the diameter $7$ of chain aggregation tree, which further enables the quicker spread of model updates in the network.
Therefore, our proposed system optimization approach effectively improve the performance of decentralized satellite neuromorphic learning.

\section{Conclusion}
\label{conclusion section}
In this paper, we developed a novel decentralized satellite neuromorphic learning framework for Space-CPN.
On the one hand, take the energy bottleneck of on-board power systems into consideration, spiking neural networks are employed to serve as the on-board intelligent models for energy-efficient data processing.
On the other hand, targeting at the performance of on-board training for such models, RelaySum is adopted to realize inter-plane model aggregation, which effectively mitigates the communication overheads of All Reduce and the degraded convergence of Gossip Averaging.
The theoretical analysis of the convergence behavior for the proposed algorithm reveals a sublinear convergence rate, which is related to the network diameter.
We then conducted system optimization to minimize the diameter of inter-plane routing tree for model aggregation on the formulated inter-plane connectivity topology.
The experimental results confirmed that our proposed framework effectively improve the energy and communication efficiency for decentralized satellite learning.

\begin{appendices}
\section{Proof of Lemma 1}
\label{lemma 1 proof}
As for the first term, we have
\begin{small}
\begin{align*}
    &\mathbb{E}\norm{\bm{\pi}^{\transpose}\tilde{\bm{W}}\bracket{\bm{G}^{t}-\mathbb{E}\bm{G}^{t}}}^{2} \\
    \leq&\mathbb{E}\norm{\frac{\pi_{0}}{NK}\sum_{i=1}^{N}\sum_{j=1}^{N}\sum_{r=0}^{R-1}\sum_{k=1}^{K}\sum_{e=0}^{E-1}\bracket{\nabla F_{i,k} - \nabla f_{i,k}}(\bm{x}_{i,k}^{t_{ij},r,e})}^{2} \\
    \stackrel{(a)}{\leq}&\frac{R^{2}E^{2}\pi_{0}^{2}}{N}\sum_{i=1}^{N}\mathbb{E}\Bigg\|\sum_{j=1}^{N}\frac{1}{R}\sum_{r=0}^{R-1}\frac{1}{K}\sum_{k=1}^{K}\frac{1}{E}\sum_{e=0}^{E-1}\Big(\nabla F_{i,k}(\bm{x}_{i,k}^{t_{ij},r,e}) \\
    &- \nabla f_{i,k}(\bm{x}_{i,k}^{t_{ij},r,e})\Big)\Bigg\|_{2}^{2} \\
    \stackrel{(b)}{\leq}&\frac{RE\pi_{0}^{2}}{NK}\sum_{i=1}^{N}\sum_{j=1}^{N}\sum_{r=0}^{R-1}\sum_{k=1}^{K}\sum_{e=0}^{E-1}\mathbb{E}\norm{\bracket{\nabla F_{i,k} - \nabla f_{i,k}}(\bm{x}_{i,k}^{t_{ij},r,e})}^{2} \\
    \leq& NR^{2}E^{2}\pi_{0}^{2}\sigma^{2},
\end{align*}
\end{small}
where $(a)$ and $(b)$ are due to Jensen's inequality, and we use $\bracket{\nabla F_{i,k} - \nabla f_{i,k}}(\bm{x})$ to serve as the abbreviation of $\nabla F_{i,k}(\bm{x}) - \nabla f_{i,k}(\bm{x})$.
As for the second term, we have
\begin{small}
\begin{align*}
    &\mathbb{E}\Fnorm{\tilde{\bm{W}}\bracket{\bm{G}^{t}-\mathbb{E}\bm{G}^{t}}}^{2}\\
    \stackrel{(a)}{\leq}& \sum_{i=1}^{N}\mathbb{E}\norm{\frac{1}{N}\sum_{j=1}^{N}\sum_{r=0}^{R-1}\sum_{k=1}^{K}\sum_{e=0}^{E-1}\bracket{\nabla F_{i,k} - \nabla f_{i,k}}(\bm{x}_{i,k}^{t_{ij},r,e})}^{2}  \\
    \stackrel{(b)}{\leq}&\frac{RE}{N^{2}K}\sum_{i=1}^{N}\sum_{j=1}^{N}\sum_{r=0}^{R-1}\sum_{k=1}^{K}\sum_{e=0}^{E-1}\mathbb{E}\norm{\bracket{\nabla F_{i,k} - \nabla f_{i,k}}(\bm{x}_{i,k}^{t_{ij},r,e})}^{2} \\
    \leq& R^{2}E^{2}\sigma^{2},
\end{align*}
\end{small}
where $(a)$ comes from Jensen's inequality and $(b)$ holds due to the independence of gradients. 

\section{Proof of Lemma 2}
\label{lemma 2 proof}
Based on the reformulated model updates given in~\eqref{simplified model updates} and Assumption 1, we have:
\begin{small}
\begin{align*}
    &\mathbb{E}f(\overline{\bm{x}}^{t+1}) \\
    =& \mathbb{E}\bracket{f(\overline{\bm{x}}^{t})-\eta\bm{\pi}^{\transpose}\tilde{\bm{W}}\bm{G}^{t}} \\
    \leq& f(\overline{\bm{x}}^{t}) - \eta\underbrace{\Big\langle \nabla f(\overline{\bm{x}}^{t}), \bm{\pi}^{\transpose}\tilde{\bm{W}}\bm{G}^{t}\Big\rangle}_{T_{1}} + \frac{\eta^{2}L}{2}\underbrace{\mathbb{E}\norm{\bm{\pi}^{\transpose}\tilde{\bm{W}}\bm{G}^{t}}^{2}}_{T_{2}}.
\end{align*}
\end{small}

As for $T_{1}$, we have:
\begin{small}
\begin{align}
    &T_{1} \notag\\
    =& NRE\pi_{0}\Big\langle \nabla f(\overline{\bm{x}}^{t}), \frac{1}{NRE\pi_{0}}\bm{\pi}^{\transpose}\tilde{\bm{W}}\bm{G}^{t}\Big\rangle \notag\\
    =& NRE\pi_{0}\Bigg(\norm{\nabla f(\overline{\bm{x}}^{t})}^{2} + \Big\langle \nabla f(\overline{\bm{x}}^{t}), \frac{1}{NRE\pi_{0}}\bm{\pi}^{\transpose}\tilde{\bm{W}}\bm{G}^{t} \notag\\ 
    &- \nabla f(\overline{\bm{x}}^{t})\Big\rangle\Bigg) \notag\\
    \stackrel{(a)}{\ge}& NRE\pi_{0} \bracket{\frac{1}{2}e_{t} - \frac{1}{2}\norm{\frac{1}{NRE\pi_{0}}\bm{\pi}^{\transpose}\tilde{\bm{W}}\bm{G}^{t} - \nabla f(\overline{\bm{x}}^{t})}^{2}} \notag\\
    \stackrel{(b)}{\ge}& NRE\pi_{0}\Bigg(\frac{1}{2}e_{t} - \frac{1}{2N^{2}}\sum_{i=1}^{N}\sum_{j=1}^{N} \tilde{T}_{1} \Bigg)
    \label{T1}
\end{align}
\end{small}
with
\begin{small}
\begin{align*}
    \tilde{T}_{1} = \norm{\frac{1}{RKE}\sum_{r=0}^{R-1}\sum_{k=1}^{K}\sum_{e=0}^{E-1}\nabla f_{j,k}(\bm{x}_{j,k}^{t_{ij},r,e}) - \nabla {f}_{j}(\overline{\bm{x}}^{t})}^{2},
\end{align*}
\end{small}
where $(a)$ holds due to $a^{2} - \langle a,b \rangle \ge \frac{a^{2}}{2} + \frac{b^{2}}{2}$ for $a,b > 0$ and $(b)$ comes from Jensen's inequality.

Then we proceed on finding the upper bound of $\tilde{T}_{1}$:
\begin{small}
\begin{align}
    &\tilde{T}_{1} \notag\\
    =& \norm{\frac{1}{RKE}\sum_{r=0}^{R-1}\sum_{k=1}^{K}\sum_{e=0}^{E-1}\nabla f_{j,k}(\bm{x}_{j,k}^{t_{ij},r,e}) - \nabla {f}_{j}(\overline{\bm{x}}^{t})}^{2} \notag\\
    =& \Bigg\|\frac{1}{RKE}\sum_{r=0}^{R-1}\sum_{k=1}^{K}\sum_{e=0}^{E-1}\nabla f_{j,k}(\bm{x}_{j,k}^{t_{ij},r,e}) \pm \frac{1}{R}\sum_{r=0}^{R-1}\nabla f_{j,k}(\bm{x}_{j}^{t_{ij},r}) \notag\\
    &\pm \nabla f(\bm{x}_{j}^{t_{ij}}) - \nabla {f}_{j}(\overline{\bm{x}}^{t})\Bigg\|_{2}^{2} \notag\\
    \leq& 3\Bigg[\norm{\frac{1}{RKE}\sum_{r=0}^{R-1}\sum_{k=1}^{K}\sum_{e=0}^{E-1}\nabla f_{j,k}(\bm{x}_{j,k}^{t_{ij},r,e}) - \frac{1}{R}\sum_{r=0}^{R-1}\nabla f_{j,k}(\bm{x}_{j}^{t_{ij},r})}^{2} \notag\\
    &+ \norm{\frac{1}{R}\sum_{r=0}^{R-1}\nabla f_{j,k}(\bm{x}_{j}^{t_{ij},r}) - \nabla f(\bm{x}_{j}^{t_{ij}})}^{2} \notag\\
    &+ \norm{\nabla f(\bm{x}_{j}^{t_{ij}}) - \nabla {f}_{j}(\overline{\bm{x}}^{t})}^{2}\Bigg] \notag\\
    \leq& 3L^{2}\Bigg[\underbrace{\frac{1}{RKE}\sum_{r=0}^{R-1}\sum_{k=1}^{K}\sum_{e=0}^{E-1}\norm{\bm{x}_{j,k}^{t_{ij},r,e}-\bm{x}_{j}^{t_{ij},r}}^{2}}_{T_{11}}\notag\\
    &+\underbrace{\frac{1}{R}\sum_{r=0}^{R-1}\norm{\bm{x}_{j}^{t_{ij},r}-\bm{x}_{j}^{t_{ij}}}^{2}}_{T_{12}} +\norm{\bm{x}_{j}^{t_{ij}}-\overline{\bm{x}}^{t}}^{2}\Bigg],
    \label{tilde T1}
\end{align}
\end{small}
where the last in equality holds due to the $L$-smoothness assumption. 
As for $T_{11}$, we have:
\begin{small}
\begin{align*}
    &T_{11} \\
    =& \frac{\eta^{2}}{RKE}\sum_{r=0}^{R-1}\sum_{k=1}^{K}\sum_{e=0}^{E-1}\norm{\sum_{l=0}^{e-1}\nabla F_{j,k}(\bm{x}_{j,k}^{t_{ij},r,l})}^{2}\\
    \leq& \frac{\eta^{2}e}{RKE}\sum_{r=0}^{R-1}\sum_{k=1}^{K}\sum_{e=0}^{E-1}\sum_{l=0}^{e-1}\Bigg\|\nabla F_{j,k}(\bm{x}_{j,k}^{t_{ij},r,l})\pm \nabla f_{j,k}(\bm{x}_{j,k}^{t_{ij},r,l}) \\
    &\pm \nabla f_{j,k}(\bm{x}_{j}^{t_{ij},r}) \pm \nabla f_{j}(\bm{x}_{j}^{t_{ij},r}) \\
    &\pm \nabla f_{j}(\bm{x}_{j}^{t_{ij}}) \pm \nabla f_{j}(\overline{\bm{x}}^{t}) \pm \nabla f(\overline{\bm{x}}^{t})\Bigg\|_{2}^{2} \\
    \leq& \frac{7\eta^{2}e}{RKE}\sum_{r=0}^{R-1}\sum_{k=1}^{K}\sum_{e=0}^{E-1}\sum_{l=0}^{e-1}\Bigg[\norm{\nabla F_{j,k}(\bm{x}_{j,k}^{t_{ij},r,l}) - \nabla f_{j,k}(\bm{x}_{j,k}^{t_{ij},r,l})}^{2} \\
    &+\norm{\nabla f_{j,k}(\bm{x}_{j,k}^{t_{ij},r,l})-\nabla f_{j,k}(\bm{x}_{j}^{t_{ij},r})}^{2} \\
    &+ \norm{\nabla f_{j,k}(\bm{x}_{j}^{t_{ij},r}) - \nabla f_{j}(\bm{x}_{j}^{t_{ij},r})}^{2} \\
    &+ \norm{\nabla f_{j}(\bm{x}_{j}^{t_{ij},r}) - \nabla f_{j}(\bm{x}_{j}^{t_{ij}})}^{2} + \norm{\nabla f_{j}(\bm{x}_{j}^{t_{ij}}) - \nabla f_{j}(\overline{\bm{x}}^{t})}^{2} \\
    &+\norm{\nabla f_{j}(\overline{\bm{x}}^{t}) - \nabla f(\overline{\bm{x}}^{t})}^{2} + \norm{\nabla f(\overline{\bm{x}}^{t})}^{2} \Bigg]\\
    \stackrel{(a)}{\leq}& \frac{7\eta^{2}e}{RKE}\sum_{r=0}^{R-1}\sum_{k=1}^{K}\sum_{e=0}^{E-1}\sum_{l=0}^{e-1}\bigg[\sigma^{2}+L^{2}\norm{\bm{x}_{j,k}^{t_{ij},r,l}-\bm{x}_{j}^{t_{ij},r}}^{2}+\delta^{2}\\
    &+L^{2}\norm{\bm{x}_{j}^{t_{ij},r}-\bm{x}_{j}^{t_{ij}}}^{2}+L^{2}\norm{\bm{x}_{j}^{t_{ij}} - \overline{\bm{x}}^{t}}^{2}+\zeta^{2} + \norm{\nabla f(\overline{\bm{x}}^{t})}^{2}\bigg] \\
    \leq& 7\eta^{2}E(E-1)L^{2}T_{11} + 7\eta^{2}E(E-1)L^{2}T_{12} \\
    &+ 7\eta^{2}E(E-1)L^{2}\mbracket{\norm{\bm{x}_{j}^{t_{ij}} - \overline{\bm{x}}^{t}}^{2}+\frac{e_{t}+z^{2}}{L^{2}}},
\end{align*}
\end{small}
where $(a)$ holds due to Assumption 1-4.
As for $T_{12}$, we have:
\begin{small}
\begin{align*}
    &T_{12} \\
    =& \frac{1}{R}\sum_{r=0}^{R-1}\norm{\bm{x}_{j}^{t_{ij},r} - \bm{x}_{j}^{t_{ij}}}^{2}\\
    \leq&\frac{\eta^{2}E^{2}r}{RKE}\sum_{r=0}^{R-1}\sum_{m=0}^{r-1}\sum_{k=1}^{K}\sum_{e=0}^{E-1}\norm{\nabla F_{j,k}(\bm{x}_{j,k}^{t_{ij},m,e})}^{2}\\
    \leq&\frac{\eta^{2}E^{2}r}{RKE}\sum_{r=0}^{R-1}\sum_{m=0}^{r-1}\sum_{k=1}^{K}\sum_{e=0}^{E-1}\Bigg\|\nabla F_{j,k}(\bm{x}_{j,k}^{t_{ij},m,e})\pm \nabla f_{j,k}(\bm{x}_{j,k}^{t_{ij},m,e}) \\
    &\pm \nabla f_{j,k}(\bm{x}_{j}^{t_{ij},m}) \pm \nabla f_{j}(\bm{x}_{j}^{t_{ij},m}) \\
    &\pm \nabla f_{j}(\bm{x}_{j}^{t_{ij}}) \pm \nabla f_{j}(\overline{\bm{x}}^{t}) \pm \nabla f(\overline{\bm{x}}^{t})\Bigg\|_{2}^{2} \\
    \leq&\frac{7\eta^{2}E^{2}r}{RKE}\sum_{r=0}^{R-1}\sum_{m=0}^{r-1}\sum_{k=1}^{K}\sum_{e=0}^{E-1}\Bigg[\norm{\nabla F_{j,k}(\bm{x}_{j,k}^{t_{ij},m,e}) - \nabla f_{j,k}(\bm{x}_{j,k}^{t_{ij},m,e})}^{2} \\
    &+\norm{\nabla f_{j,k}(\bm{x}_{j,k}^{t_{ij},m,e})-\nabla f_{j,k}(\bm{x}_{j}^{t_{ij},m})}^{2} \\
    &+ \norm{\nabla f_{j,k}(\bm{x}_{j}^{t_{ij},m}) - \nabla f_{j}(\bm{x}_{j}^{t_{ij},m})}^{2} \\
    &+ \norm{\nabla f_{j}(\bm{x}_{j}^{t_{ij},m}) - \nabla f_{j}(\bm{x}_{j}^{t_{ij}})}^{2} + \norm{\nabla f_{j}(\bm{x}_{j}^{t_{ij}}) - \nabla f_{j}(\overline{\bm{x}}^{t})}^{2} \\
    &+\norm{\nabla f_{j}(\overline{\bm{x}}^{t}) - \nabla f(\overline{\bm{x}}^{t})}^{2} + \norm{\nabla f(\overline{\bm{x}}^{t})}^{2} \Bigg]\\
    \stackrel{(a)}{\leq}& \frac{7\eta^{2}E^{2}r}{RKE}\sum_{r=0}^{R-1}\sum_{m=0}^{r-1}\sum_{k=1}^{K}\sum_{e=0}^{E-1}\bigg[\sigma^{2}+L^{2}\norm{\bm{x}_{j,k}^{t_{ij},m,e}-\bm{x}_{j}^{t_{ij},m}}^{2}+\delta^{2}\\
    &+L^{2}\norm{\bm{x}_{j}^{t_{ij},m}-\bm{x}_{j}^{t_{ij}}}^{2}+L^{2}\norm{\bm{x}_{j}^{t_{ij}} - \overline{\bm{x}}^{t}}^{2}+\zeta^{2} + \norm{\nabla f(\overline{\bm{x}}^{t})}^{2}\bigg] \\
    \leq& 7\eta^{2}E^{2}R(R-1)L^{2}T_{11}+7\eta^{2}E^{2}R(R-1)L^{2}T_{12}\\
    &+7\eta^{2}E^{2}R(R-1)L^{2}\mbracket{\norm{\bm{x}_{j}^{t_{ij}} - \overline{\bm{x}}^{t}}^{2}+\frac{e_{t}+z^{2}}{L^{2}}},
\end{align*}
\end{small}
where $(a)$ is due to Assumption 1-4.
Combine the upper bound of $T_{11}$ and $T_{12}$, we can obtain:
\begin{small}
\begin{align*}
    \begin{cases}
        T_{11} &\leq A_{1}T_{12} + B_{1}\\
        T_{12} &\leq A_{2}T_{12} + B_{2}
    \end{cases}
\end{align*}
\end{small}
where $A_{1} = \frac{D_{1}}{1-D_{1}}$, $A_{2} = \frac{D_{2}}{1-D_{2}}$, $B_{1}=\frac{D_{1}}{1-D_{1}}(\|\bm{x}_{j}^{t_{ij}} -\overline{\bm{x}}^{t}\|_{2}^{2}+\frac{e_{t}+z^{2}}{L^{2}})$, and $B_{2}=\frac{D_{2}}{1-D_{2}}(L^{2}\|\bm{x}_{j}^{t_{ij}} -\overline{\bm{x}}^{t}\|_{2}^{2}+\frac{e_{t}+z^{2}}{L^{2}})$.
This further leads to:
\begin{small}
\begin{align*}
     T_{11} + T_{12} \leq \frac{1+A_{2}}{1-A_{1}A_{2}}B_{1} + \frac{1+A_{1}}{1-A_{1}A_{2}}B_{2},
\end{align*}
\end{small}
by setting $A_{1}\leq\frac{1}{2}$ and $A_{2}\leq\frac{1}{2}$, we have
\begin{small}
\begin{align}
    T_{11} + T_{12} \leq& \frac{1+A_{2}}{1-A_{1}A_{2}}B_{1} + \frac{1+A_{1}}{1-A_{1}A_{2}}B_{2}\notag\\
    \leq& \bracket{\frac{2D_{1}}{1-D_{1}}+\frac{2D_{2}}{1-D_{2}}}\bracket{\norm{\bm{x}_{j}^{t_{ij}} - \overline{\bm{x}}^{t}}^{2}+\frac{e_{t}+z^{2}}{L^{2}}} \notag\\
    \leq& 3\bracket{D_{1}+D_{2}}\bracket{\norm{\bm{x}_{j}^{t_{ij}} - \overline{\bm{x}}^{t}}^{2}+\frac{e_{t}+z^{2}}{L^{2}}}
    \label{T11+T12}
\end{align}
\end{small}
with $\eta \leq \frac{\tilde{\pi}_{0}}{36REL} \leq \frac{1}{36REL}$.
Substitute~\eqref{T11+T12} in~\eqref{tilde T1}, we have:
\begin{small}
\begin{align}
    &\tilde{T}_{1} \notag\\
    \leq& 3L^{2}\bracket{T_{11}+T_{12}+\norm{\bm{x}_{j}^{t_{ij}}-\overline{\bm{x}}^{t}}^{2}} \notag\\
    \leq& \bracket{9D_{1}+9D_{2}}e_{t} + \bracket{9D_{1}+9D_{2}+3}L^{2}\norm{\bm{x}_{j}^{t_{ij}}-\overline{\bm{x}}^{t}}^{2} \notag\\
    &+ \bracket{9D_{1}+9D_{2}}z^{2}.
    \label{bound of tilde T1}
\end{align}
\end{small}
Then by substituting~\eqref{tilde T1} into~\eqref{T1}, we can obtain the bound of $T_{1}$:
\begin{small}
\begin{align}
    &T_{1} \notag\\
    \ge& NRE\pi_{0}\bigg(\frac{1}{2}e_{t}-\frac{9D_{1}+9D_{2}}{2}e_{t}-\frac{9D_{1}+9D_{2}}{2}z^{2} \notag\\
    &-\frac{9D_{1}+9D_{2}+3}{2N^{2}}L^{2}\sum_{i=1}^{N}\sum_{j=1}^{N}\norm{\bm{x}_{j}^{t_{ij}}-\overline{\bm{x}}^{t}}^{2}\bigg) \notag\\
    \ge& NRE\pi_{0}\bigg(\frac{1-9D_{1}-9D_{2}}{2}e_{t} - \frac{9D_{1}+9D_{2}}{2}z^{2} \notag\\
    &- \frac{9D_{1}+9D_{2}+3}{2}L^{2}\theta_{t}\bigg)
\end{align}
\end{small}

As for $T_{2}$, we have:
\begin{small}
\begin{align}
    &T_{2} \notag\\
    \leq& 2\mathbb{E}\norm{\bm{\pi}^{\transpose}\tilde{\bm{W}}\bm{G}^{t}-\bm{\pi}^{\transpose}\tilde{\bm{W}}\mathbb{E}\bm{G}^{t}}^{2} + 2R^{2}E^{2}\norm{\frac{1}{RE}\bm{\pi}^{\transpose}\tilde{\bm{W}}\mathbb{E}\bm{G}^{t}}^{2} \notag\\
    \leq& 2NR^{2}E^{2}\pi_{0}^{2}\sigma^{2}+ 4N^{2}R^{2}E^{2}\pi_{0}^{2}\norm{\nabla f(\overline{\bm{x}}^{t})}^{2} +4R^{2}E^{2}\pi_{0}^{2}\notag\\
    &\sum_{i=1}^{N}\sum_{j=1}^{N}\norm{\frac{1}{RKE}\sum_{r=0}^{R-1}\sum_{k=1}^{K}\sum_{e=0}^{E-1}\nabla f_{j,k}(\bm{x}_{j,k}^{t_{ij},r,e}) - \nabla {f}_{j}(\overline{\bm{x}}^{t})}^{2} \notag\\
    \stackrel{(a)}{\leq}& 2NR^{2}E^{2}\pi_{0}^{2}\sigma^{2} + \bracket{36D_{1}+36D_{2}+4}N^{2}R^{2}E^{2}\pi_{0}^{2}e_{t} \notag\\
    &+\bracket{36D_{1}+36D_{2}+12}N^{2}R^{2}E^{2}\pi_{0}^{2}L^{2}\theta_{t}\notag\\
    &+\bracket{36D_{1}+36D_{2}}N^{2}R^{2}E^{2}\pi_{0}^{2}z^{2},
\end{align}
\end{small}
where $(a)$ holds due to the bound of $\tilde{T}_{1}$ we have derived as in~\eqref{bound of tilde T1}.
Combine the bounds of $T_{1}$ and $T_{2}$, we have the following result:
\begin{small}
\begin{align*}
    &r_{t+1} \\
    \leq& r_{t} - \eta T_{1} + \frac{\eta^{2}L}{2}T_{2} \\
    \leq& r_{t} - NRE\pi_{0}\eta\bigg(\frac{1-9D_{1}-9D_{2}}{2}e_{t} - \frac{9D_{1}+9D_{2}}{2}z^{2} \\
    &- \frac{9D_{1}+9D_{2}+3}{2}L^{2}\theta_{t}\bigg) + NR^{2}E^{2}\pi_{0}^{2}\eta^{2}L\sigma^{2} \\
    &+ \bracket{18D_{1}+18D_{2}+2}N^{2}R^{2}E^{2}\pi_{0}^{2}\eta^{2}Le_{t} \\
    &+\bracket{18D_{1}+18D_{2}+6}N^{2}R^{2}E^{2}\pi_{0}^{2}\eta^{2}L^{3}\theta_{t}\\
    &+\bracket{18D_{1}+18D_{2}}N^{2}R^{2}E^{2}\pi_{0}^{2}\eta^{2}Lz^{2} \\
    \leq& r_{t} - NRE\pi_{0}\eta\Bigg[\bigg(\frac{1-9D_{1}-9D_{2}}{2} - NRE\pi_{0}\eta L \\
    &\cdot(18D_{1}+18D_{2}+2)\bigg)e_{t} + \bigg(\frac{9D_{1}+9D_{2}+3}{2}+NRE\pi_{0}\eta L \\
    &\cdot(18D_{1}+18D_{2}+6)\bigg)L^{2}\theta_{t} + \bigg(\frac{9D_{1}+9D_{2}}{2} + NRE\pi_{0}\eta L\\
    &\cdot(18D_{1}+18D_{2})\bigg)z^{2} + NR^{2}E^{2}\pi_{0}^{2}\eta^{2}L\sigma^{2}\Bigg]
\end{align*}
\end{small}

The setting of learning rate guarantee that $\eta \leq \frac{1}{36NRE\pi_{0}L} $ and $\eta \leq \frac{1}{36NREL}$.
Based on this we have: 
\begin{small}
\begin{align*}
    D_{1}+D_{2} =& 7 \bracket{\eta^{2}E(E-1)L^{2}+\eta^{2}E^{2}R(R-1)L^{2}} \\
    \leq& 7 \bracket{\frac{E(E-1)L^{2}}{1296N^{2}R^{2}E^{2}L^{2}}+\frac{E^{2}R(R-1)L^{2}}{1296N^{2}R^{2}E^{2}L^{2}}} \\
    \leq& \frac{14}{1296} = \frac{7}{648},
\end{align*}
\end{small}
which further leads to:
\begin{small}
\begin{align*}
    \begin{cases}
    &\frac{1-9D_{1}-9D_{2}}{2} - NRE\pi_{0}\eta L(18D_{1}+18D_{2}+2) \ge \frac{1}{2} \\
    &\frac{9D_{1}+9D_{2}+3}{2}+NRE\pi_{0}\eta L(18D_{1}+18D_{2}+6) \leq 2 \\
    &\frac{9D_{1}+9D_{2}}{2} + NRE\pi_{0}\eta L(18D_{1}+18D_{2}) \leq 5(D_{1}+5D_{2})
\end{cases}
\end{align*}
\end{small}
Then we can reach the final result of the decrease of the non-convex objective:
\begin{small}
\begin{align*}
        r_{t+1} \leq &  r_{t} - \frac{NRE\pi_{0}\eta}{4}e_{t}+2NRE\pi_{0}\eta L^{2}\theta_{t}+NR^{2}E^{2}\pi_{0}^{2}\eta^{2}L\sigma^{2}\\
        &+5\bracket{D_{1}+D_{2}}NRE\pi_{0}\eta z^{2}.
\end{align*}
\end{small}

\section{Proof of Lemma 4}
\label{lemma 4 proof}
Based on the Lemma 18 in~\cite{vogels2021relaysum} with respect to consensus and our analysis of $\tilde{T}_{1}$ in Lemma 2, we can reach the following result:
\begin{small}
\begin{align}
    \frac{1}{T}\sum_{t=0}^{T-1}\theta_{t} \leq& (1-\frac{q}{2})^{2m}\frac{1}{T}\sum_{t=0}^{T}\theta_{t}+\frac{4C_{1}^{2}m^{2}\eta^{2}}{qT}\sum_{t=0}^{T-1}\bigg[R^{2}E^{2} \notag\\
    &\cdot\Big((9D_{1}+9D_{2}+3)L^{2}\theta_{t}+(9D_{1}+9D_{2}+1)e_{t} \notag\\
    &+(9D_{1}+9D_{2})z^{2}\Big)\bigg]+2C_{1}^{2}m^{2}R^{2}E^{2}\frac{\sigma^{2}}{N}.
    \label{consensus proof 1}
\end{align}
\end{small}
By setting $\eta \leq \frac{q}{36C_{1}mREL}$, we have:
\begin{small}
\begin{align*}
    \frac{4C_{1}^{2}m^{2}}{q}R^{2}E^{2}(9D_{1}+9D_{2}+3)L^{2}\eta^{2} \leq \frac{q}{4}.
\end{align*}
\end{small}
Then by rearranging terms in~\eqref{consensus proof 1}, we can obtain the result in~\eqref{bound of consensus distance}.

\section{Proof of Theorem 1}
\label{theorem 1 proof}
Rearranging the terms in~\eqref{bound of non-convex objective} and average over $t$, we have:
\begin{small}
\begin{align*}
    &\frac{1}{T}\sum_{t=0}^{T-1}e_{t} \\
    \leq& \frac{1}{T}\sum_{t=0}^{T-1}\bracket{\frac{4r_{t}}{NRE\pi_{0}\eta}-\frac{4r_{t+1}}{NRE\pi_{0}\eta}}+8L^{2}\frac{1}{T}\sum_{t=0}^{T-1}\theta_{t} \\
    &+8RE\pi_{0}L\eta\sigma^{2}+20(D_{1}+D_{2})z^{2} \\
    \leq& \frac{4r_{0}}{NRE\pi_{0}\eta T}+8L^{2}\Bigg(\frac{20C_{1}^{2}m^{2}R^{2}E^{2}\eta^{2}}{q^{2}}\frac{1}{T}\sum_{t=0}^{T-1}e_{t}\\
    &+\frac{8C_{1}^{2}m^{2}R^{2}E^{2}\eta^{2}}{Nq^{2}}\sigma^{2}+\frac{144(D_{1}+D_{2})C_{1}^{2}m^{2}R^{2}E^{2}\eta^{2}}{q^{2}}z^{2}\Bigg) \\
    &+8RE\pi_{0}L\eta\sigma^{2}+20(D_{1}+D_{2})z^{2}.
\end{align*}
\end{small}
With the constraint of learning rate that $\eta \leq \frac{q\tilde{\pi}_{0}}{36C_{1}mREL}$, we have:
\begin{small}
\begin{align*}
    8L^{2} \frac{20C_{1}^{2}m^{2}R^{2}E^{2}\eta^{2}}{q^{2}} \leq \frac{1}{2}.
\end{align*}
\end{small}
Based on this we can rearrange the terms and obtain:
\begin{small}
\begin{align*}
    &\frac{1}{T}\sum_{t=0}^{T-1}e_{t} \\
    \leq& \frac{8r_{0}}{NRE\pi_{0}\eta T} + \bracket{\frac{128C_{1}^{2}m^{2}R^{2}E^{2}L^{2}\eta^{2}}{Nq^{2}}+16RE\pi_{0}L\eta}\sigma^{2} \\
    &+7\bracket{E(E-1)+E^{2}R(R-1)}\bracket{\frac{16C_{1}^{2}m^{2}}{9q^{2}N^{2}\pi_{0}^{2}L^{2}}+40}L^{2}\eta^{2}z^{2} \\
    \leq& \frac{8}{NRE\pi_{0}}\frac{r_{0}}{\eta T} + 16RE\pi_{0}L\sigma^{2}\eta + \Bigg[\frac{128C_{1}^{2}m^{2}R^{2}E^{2}L^{2}}{Nq^{2}} \\
    &+ 7\bracket{E(E-1)+E^{2}R(R-1)}\bracket{\frac{16C_{1}^{2}m^{2}}{9q^{2}N^{2}\pi_{0}^{2}L^{2}}+40}L^{2}z^{2} \Bigg]\eta^{2} \\
    \stackrel{(a)}{\leq}& \frac{8}{NRE\pi_{0}}\Bigg\{2\bracket{2NR^{2}E^{2}\pi_{0}^{2}L\sigma^{2}\frac{r_{0}}{T}}^{\frac{1}{2}}+2\Bigg[NRE\pi_{0}\\
    &\cdot\bigg(\frac{16C_{1}^{2}m^{2}R^{2}E^{2}L^{2}\eta^{2}}{Nq^{2}}+7\bracket{E(E-1)+E^{2}R(R-1)}\\
    &\cdot\bracket{\frac{2C_{1}^{2}m^{2}}{9q^{2}N^{2}\pi_{0}^{2}L^{2}}+5}L^{2}z^{2}\bigg)\Bigg]^{\frac{1}{3}}\bracket{\frac{r_{0}}{T}}^{\frac{2}{3}}+\frac{36C_{1}mREL}{q\tilde{\pi}_{0}}\frac{r_{0}}{T}\Bigg\} \\
    \leq& 16\bracket{\frac{2L\sigma^{2}r_{0}}{NT}}^{\frac{1}{2}}+16\bracket{\frac{4C\sqrt{\tilde{\tau}}L\sigma r_{0}}{\rho\sqrt{N}T}}^{\frac{2}{3}}+\frac{288CL\sqrt{\tilde{\tau}}r_{0}}{\rho T}\\
    &+16\mbracket{\frac{\sqrt{7E(E-1)+7E^{2}R(R-1)}}{NRE\pi_{0}}\sqrt{\frac{2C^{2}\tilde{\tau}}{9\rho^{2}L^{2}}+5}\frac{Lzr_{0}}{T}}^{\frac{2}{3}},
\end{align*}
\end{small}
where $(a)$ holds due to Lemma 15 in~\cite{koloskova2020unified}.

\end{appendices}

\bibliographystyle{ieeetr}
\bibliography{ref} 
\end{document}